\documentclass{article}

\usepackage[accepted]{icml2023}

\usepackage[utf8]{inputenc} 
\usepackage[T1]{fontenc}    
\usepackage{booktabs}       
\usepackage{amsfonts}       
\usepackage{nicefrac}       
\usepackage{amsmath}
\usepackage{amssymb}
\usepackage{mathtools}
\usepackage{amsthm}
\usepackage{microtype}      
\usepackage{xcolor}         
\usepackage{xspace}         
\usepackage{subcaption}
\usepackage{hyperref}
\usepackage{url}
\usepackage{multirow}
\usepackage{enumitem}
\usepackage{wrapfig}
\usepackage{array}
\usepackage{balance}

\newcolumntype{H}{>{\setbox0=\hbox\bgroup}c<{\egroup}@{}}

\usepackage{amsmath,amsfonts,bm,dsfont,diagbox,mathtools}
\usepackage{letltxmacro}

\makeatletter

\LetLtxMacro\orgvdots\vdots
\LetLtxMacro\orgddots\ddots

\DeclareRobustCommand\vdots{
  \mathpalette\@vdots{}
}
\newcommand*{\@vdots}[2]{
  \sbox0{$#1\cdotp\cdotp\cdotp\m@th$}
  \sbox2{$#1.\m@th$}
  \vbox{
    \dimen@=\wd0 
    \advance\dimen@ -3\ht2 
    \kern.5\dimen@
    \dimen@=\wd2 
    \advance\dimen@ -\ht2 
    \dimen2=\wd0 
    \advance\dimen2 -\dimen@
    \vbox to \dimen2{
      \offinterlineskip
      \copy2 \vfill\copy2 \vfill\copy2 
    }
  }
}
\DeclareRobustCommand\ddots{
  \mathinner{
    \mathpalette\@ddots{}
    \mkern\thinmuskip
  }
}
\newcommand*{\@ddots}[2]{
  \sbox0{$#1\cdotp\cdotp\cdotp\m@th$}
  \sbox2{$#1.\m@th$}
  \vbox{
    \dimen@=\wd0 
    \advance\dimen@ -3\ht2 
    \kern.5\dimen@
    \dimen@=\wd2 
    \advance\dimen@ -\ht2 
    \dimen2=\wd0 
    \advance\dimen2 -\dimen@
    \vbox to \dimen2{
      \offinterlineskip
      \hbox{$#1\mathpunct{.}\m@th$}
      \vfill
      \hbox{$#1\mathpunct{\kern\wd2}\mathpunct{.}\m@th$}
      \vfill
      \hbox{$#1\mathpunct{\kern\wd2}\mathpunct{\kern\wd2}\mathpunct{.}\m@th$}
    }
  }
}

\let\save@mathaccent\mathaccent
\newcommand*\if@single[3]{
    \setbox0\hbox{${\mathaccent"0362{#1}}^H$}
    \setbox2\hbox{${\mathaccent"0362{\kern0pt#1}}^H$}
    \ifdim\ht0=\ht2 #3\else #2\fi
    }
\newcommand*\rel@kern[1]{\kern#1\dimexpr\macc@kerna}
\newcommand*\widebar[1]{{\@ifnextchar^{{\wide@bar{#1}{0}}}{\wide@bar{#1}{1}}}}
\newcommand*\wide@bar[2]{\if@single{#1}{\wide@bar@{#1}{#2}{1}}{\wide@bar@{#1}{#2}{2}}}
\newcommand*\wide@bar@[3]{
\begingroup
\def\mathaccent##1##2{
    \let\mathaccent\save@mathaccent
    \if#32 \let\macc@nucleus\first@char \fi
    \setbox\z@\hbox{$\macc@style{\macc@nucleus}_{}$}
    \setbox\tw@\hbox{$\macc@style{\macc@nucleus}{}_{}$}
    \dimen@\wd\tw@
    \advance\dimen@-\wd\z@
    \divide\dimen@ 3
    \@tempdima\wd\tw@
    \advance\@tempdima-\scriptspace
    \divide\@tempdima 10
    \advance\dimen@-\@tempdima
    \ifdim\dimen@>\z@ \dimen@0pt\fi
    \rel@kern{0.6}\kern-\dimen@
    \if#31
        \overline{\rel@kern{-0.6}\kern\dimen@\macc@nucleus\rel@kern{0.4}\kern\dimen@}
        \advance\dimen@0.4\dimexpr\macc@kerna
        \let\final@kern#2
        \ifdim\dimen@<\z@ \let\final@kern1\fi
        \if\final@kern1 \kern-\dimen@\fi
    \else
        \overline{\rel@kern{-0.6}\kern\dimen@#1}
    \fi
}
\macc@depth\@ne
\let\math@bgroup\@empty \let\math@egroup\macc@set@skewchar
\mathsurround\z@ \frozen@everymath{\mathgroup\macc@group\relax}
\macc@set@skewchar\relax
\let\mathaccentV\macc@nested@a
\if#31
    \macc@nested@a\relax111{#1}
\else
    \def\gobble@till@marker##1\endmarker{}
    \futurelet\first@char\gobble@till@marker#1\endmarker
    \ifcat\noexpand\first@char A\else
        \def\first@char{}
    \fi
    \macc@nested@a\relax111{\first@char}
\fi
\endgroup
}
\makeatother

\newtheorem{definition}{Definition}

\def\eqref#1{equation~\ref{#1}}

\def\1{\bm{1}}

\def\rmX{{\mathbf{X}}}

\def\vzero{{\bm{0}}}
\def\vone{{\bm{1}}}

\def\vb{{\bm{b}}}

\def\vo{{\bm{o}}}

\def\vx{{\bm{x}}}

\def\vz{{\bm{z}}}

\def\mW{{\bm{W}}}

\DeclareMathAlphabet{\mathsfit}{\encodingdefault}{\sfdefault}{m}{sl}
\SetMathAlphabet{\mathsfit}{bold}{\encodingdefault}{\sfdefault}{bx}{n}

\def\cT{{\mathcal{T}}}
\def\cU{{\mathcal{U}}}

\def\sR{{\mathbb{R}}}

\DeclareMathOperator*{\argmin}{arg\,min}

\newcommand{\update}[1]{#1}

\newcommand{\x}{{\rmX}}

\newcommand{\xpred}{{\hat{\x}}}

\newcommand{\T}{{T}} 
\newcommand{\tobs}{{r}}

\newcommand{\xt}[1]{\x_{t_{#1}}}

\newcommand{\method}{MetaPhysiCa\xspace}

\newcommand{\methodsindyvtwo}{\method\xspace}

\newcommand{\phy}{\text{phy}}
\newcommand{\nn}{\text{nn}}

\newcommand{\nan}{\text{NaN}^*}
\newcommand{\param}{\text{param}}

\newcommand{\lambdavrex}{\lambda_\text{REx}}
\newcommand{\lambdaone}{\lambda_\Phi}

\newcommand{\thetapsi}{{\mW}}

\newcommand{\thetaglobal}[1]{{\bm \xi_{#1}}}
\newcommand{\thetaglobalelem}[2]{{\xi_{#1, #2}}}

 \usepackage[capitalize,noabbrev]{cleveref}

\usepackage{graphicx}
\graphicspath{{./figs/}{./figs/final_plots}}

\icmltitlerunning{\method: OOD Robustness in PIML}

\begin{document}

\twocolumn[
\icmltitle{\method: OOD Robustness in Physics-informed Machine Learning}

\icmlsetsymbol{equal}{*}

\begin{icmlauthorlist}
\icmlauthor{S Chandra Mouli}{purdue}
\icmlauthor{Muhammad Ashraful Alam}{eepurdue}
\icmlauthor{Bruno Ribeiro}{purdue}
\end{icmlauthorlist}

\icmlaffiliation{purdue}{Department of Computer Science, Purdue University, Indiana, USA}
\icmlaffiliation{eepurdue}{Department of Electrical and Computer Engineering, Purdue University, Indiana, USA}

\icmlcorrespondingauthor{S Chandra Mouli}{chandr@purdue.edu}

\icmlkeywords{Machine Learning, ICML}

\vskip 0.3in
]

\printAffiliationsAndNotice{}

\newcommand{\fix}{\marginpar{FIX}}
\newcommand{\new}{\marginpar{NEW}}

\begin{abstract}

A fundamental challenge in physics-informed machine learning (PIML) is the design of robust PIML methods for out-of-distribution (OOD) forecasting tasks. These OOD tasks require learning-to-learn from observations of the same (ODE) dynamical system with different unknown ODE parameters, and demand accurate forecasts even \update{under out-of-support initial conditions and out-of-support ODE parameters}. In this work we propose a solution for such tasks, which we define as a meta-learning procedure for causal structure discovery (including invariant risk minimization). Using three different OOD tasks, we empirically observe that the proposed approach significantly outperforms existing state-of-the-art PIML and deep learning methods.

\end{abstract}

\section{Introduction}

Physics-informed machine learning (PIML) (e.g.,~\citep{willard2020,xingjian2015convolutional,Lusch2018,Yeo2019,Raissi2018b,kochkov2021machine}) seeks to combine the strengths of physics and machine learning models and has positively impacted fields as diverse as biological sciences~\citep{yazdani2020systems}, climate science~\citep{faghmous2014big}, turbulence modeling~\citep{ling2016a,wang2020}, among others. 
PIML achieves substantial success in tasks where the test data comes from the same distribution as the training data ({\em in-distribution tasks}).

Unlike the PIML works described above, this paper considers an out-of-distribution (OOD) change in the initial system state \update{and unknown parameters} of the dynamical system, possibly with different train and test distribution supports (illustrated in \cref{fig:main_meta_physics_task}(a,b)).
In this setting, we observe that existing state-of-the-art PIML models perform significantly worse than their performance in-distribution, even in PIML methods designed with OOD robustness in mind~\citep{wang2021a,kirchmeyer2022}.
This is because the standard ML part of PIML, which tends to learn spurious associations, will perform poorly in our OOD setting.
We then propose a promising solution: Combine {\em meta learning} with {\em causal structure discovery} to learn an ODE model that is robust to OOD initial conditions \update{and can adapt to OOD parameters of the dynamical system}. 
In our OOD tasks, OOD robustness means that the robustness is tied to interventions over the initial conditions and unknown parameters of the system, not on arbitrary interventions as the system evolves from the initial state.
This is an important distinction. There can be multiple ODE models that will be equally OOD robust, and robust ODEs may not correctly predict system trajectories under arbitrary system interventions (\citet{rubenstein2016deterministic} discusses the effect of arbitrary interventions in physics models).

\vspace{-10pt}
\paragraph{Contributions}
This work proposes a {\em hybrid transductive-inductive modeling approach} learning for more robust ODEs using {\em meta learning} and {\em causal structure discovery} (e.g., via $L_1$ regularization~\citep{zheng2018dags}, which can be combined with invariant risk minimization~\citep{arjovsky2019invariant,krueger2021out}).
More precisely, our contributions are:
\begin{enumerate}[leftmargin=*]
\item  We show that state-of-the-art PIML and deep learning methods fail in test examples with OOD initial conditions and/or OOD system parameters. Prior work~\citep{wang2020bridging} showed that deep learning-only methods fail in OOD tasks, and argued physics models and PIML methods would succeed, including a proposed OOD solution~\citep{wang2021a}. Here we show that PIML methods also fail (or perform poorly) OOD, including the solution in~\citet{wang2021a}.
\item {\em We proposed a hybrid transductive-inductive learning framework for ODEs via meta learning}: As in transductive methods, we will consider each training and test examples as separate tasks, but like inductive methods, the tasks are dependent and knowledge can be transferred between the learned ODEs.
By {\em meta learning} we mean the definition in \citep[Chapter 1.2]{thrun1998learning}, where given:
(a) a family of $M$ tasks (a task is a single experiment in our setting), $i=1,\ldots,M$; (b) training experience for each task $i\in \{1,\ldots,M\}$, which for us are the time series observations of an experiment $\xt{0}^{(i)},\ldots,\xt{T}^{(i)}$, and; (c) a family of performance measures (e.g., one for each task) described by the risk function $R^{(i)}$;
our algorithm will {\em meta learn} such that performance at each task improves with experience (more observations) and with the number of tasks (number of experiments).
For an algorithm to fit this definition, there must be a transfer of knowledge between multiple tasks that has a positive impact on expected task performance across all tasks.

\item {\em Learning ODEs as structural causal discovery}. In order to learn an ODE that is robust to OOD changes in initial conditions (with possibly non-overlapping training and test distribution supports), we define a family of structural causal models and perform a structural causal search in order to find the correct model for our task (which is assumed to be in the family). 
We test common structural causal discovery approaches for linear models: $\ell_1$-regularization with and without an invariant risk minimization-type objective, which we observe achieve similar empirical results.
\end{enumerate}
The proposed method is then empirically validated using three commonly-used simulated physics tasks (with measurement noise): Damped pendulum systems~\citep{guen2020}, predator-prey systems~\citep{wang2020bridging}, and epidemic modeling~\citep{wang2020bridging}.
\update{Train and test distributions of initial conditions and unknown ODE parameters have non-overlapping support.}

\section{Dynamical System Forecasting as a Meta Learning Task}

We formally describe the task of forecasting a dynamical system with a focus on the out-of-distribution scenarios.

\begin{figure*}[t]
	\centering
	\includegraphics[scale=0.75]{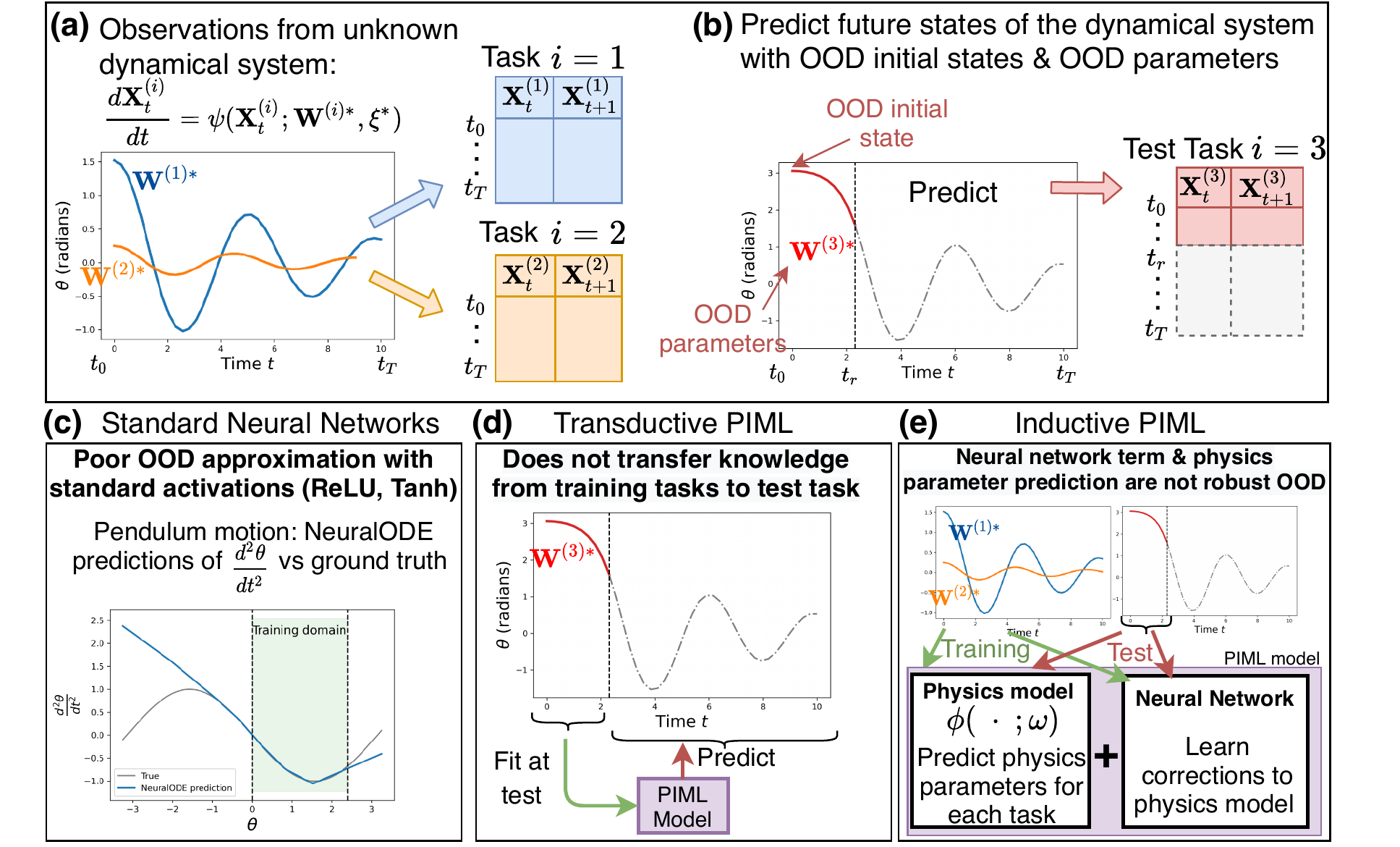}
	\caption{\small
	Dynamical system OOD problem definition and traditional approaches to address it.
	\textbf{(a)} Training data consists of multiple observations from the same dynamical system with different parameters $\thetapsi^{(i)*}$. 
	Each training curve can be seen as a different task $i$ where the goal is to predict $\x^{(i)}_{t+1}$ from $\x^{(i)}_t$ for all $t$. 
	\textbf{(b)} At test, we are given observations till $t_\tobs$ (red solid) and the goal is to predict the future observations till $t_\T$ (gray dashed).  The initial conditions and the unknown ODE parameters can be out-of-distribution in test.
	\textbf{(c)} Shows OOD failure of a standard neural network (NeuralODE~\citep{chen2018}) for dynamical system forecasting. 
	When trained to predict the motion of damped pendulum, the model predicts accurately in the training domain (green shaded), but predicts a linear function outside the training domain. 
	\textbf{(d)} Transductive PIML methods (e.g.,~\citep{Raissi,Brunton2016}) are not able to transfer knowledge from training tasks to a test task with different $\thetapsi^{*}$. 
	Thus, these models can be fit only using test observations till time $t_\tobs$ ignoring the training data.
	\textbf{(e)} Inductive PIML methods (e.g., \citep{guen2020,mehta2021}) use a known (possibly incomplete) physics model $\phi(~\cdot~; \omega)$ and inductively predict its parameters $\omega$ for each task, typically using a neural network. 
	However, predicting these physics parameters at test this way is not robust.
	Furthermore, they use a neural network term to correct for the incomplete physics model and face the same robustness issue discussed in \textbf{(c)}.
	}
	\label{fig:main_meta_physics_task}
\vspace{-10pt}
\end{figure*}

\begin{definition}[Dynamical system forecasting task] \label{def:forecasting_task}
In what follows we describe our task:
\vspace{-5pt}
\begin{enumerate}[leftmargin=*,itemsep=0pt]
    \item {\bf Training data (depicted in \cref{fig:main_meta_physics_task}(a)):} In training, we are given a set of $M$ experiments, which we will denote as $M$ {\em tasks}.
    Task $i \in \{1,\ldots,M\}$ has an associated (hidden) environment $e^{(i)}$. Different tasks can have the same environment.
    Let $\mathcal{T}^{(i)} := \xt{0}^{(i)}, \ldots, \xt{\T^{(i)}}^{(i)}$ denote the noisy observations of our dynamical system,  with $\x_t^{(i)} := \vx_t^{(i)} + {\bm \varepsilon}^{(i)}_t$, where
    \begin{equation} \label{eq:dynamical_system}
      \frac{d\vx^{(i)}_t}{dt} = \psi(\vx^{(i)}_t; \thetapsi^{(i)*}, \thetaglobal{}^*)  \:,
    \end{equation}
    $\{t_0, \ldots, t_{\T^{(i)}}\}$ are regularly-spaced discrete time steps~\footnote{Regularly spaced intervals are not strictly necessary for our method, but it makes its implementation simpler.}, $\vx^{(i)}_t \in \sR^d$ is the (hidden) state of the system at time $t$ during experiment (task) $i$, 
    ${\bm \varepsilon}^{(i)}_t$ are independent zero-mean Gaussian noises, 
    $\psi$ is an unknown deterministic function with task-dependent parameters $\thetapsi^{(i)*}$ and global task-independent parameters $\thetaglobal{}^*$, both hidden.
    
    The distribution of initial conditions $\vx^{(i)}_{t_0} \sim P(\x_{t_0} | E=e^{(i)})$ \update{and that of hidden parameters $\thetapsi^{(i)*} \sim P(\thetapsi^* | E=e^{(i)})$ for task $i$ may depend on its environment $e^{(i)}$.}
    The unknown parameters $\thetaglobal{}^*$ remain constant across environments.
    \item {\bf Test data ((depicted in \cref{fig:main_meta_physics_task}(b)):} At test, we are given noisy observations of the initial sequence $\widetilde{\mathcal{T}}^{(M+1)}:= \xt{0}^{(M+1)}, \ldots, \xt{\tobs}^{(M+1)}$, where ${\tobs}$ is generally small, of the dynamical system
    \[
    \frac{d\vx^{(M+1)}_t}{dt} = \psi(\vx^{(M+1)}_t; \thetapsi^{(M+1)*}, \thetaglobal{}^*)
    \]
    with initial condition $\vx^{(M+1)}_{t_0} \sim P(\x_{t_0} | E=e^{(M+1)})$, \update{(unknown) system parameters $\thetapsi^{(M+1)*} \sim P(\thetapsi^* | E=e^{(M+1)})$} and hidden global parameters $\thetaglobal{}^*$ the same as in training. {\bf Our task is to predict} $\xt{\tobs+1}^{(M+1)}, \ldots, \xt{\T^{(M+1)}}^{(M+1)}$ from the initial observations $\widetilde{\mathcal{T}}^{(M+1)}$, using the inductive knowledge obtained from the training data.
    \item \textbf{OOD initial conditions and system parameters:} Initial conditions in training $\{P(\x_{t_0} ~|~ E=e^{(i)})\}_{i=1}^M$, can be different from initial conditions in test $P(\x_{t_0} ~|~ E=e^{(M+1)})$ with possibly non-overlapping support due to the presence of an environment unseen in training. 
    \update{Similarly, the hidden parameters in training $\{P(\thetapsi^* ~|~ E=e^{(i)})\}_{i=1}^M$, can have different support from those in test $P(\thetapsi^* | E=e^{(M+1)})$.}
\end{enumerate}
\end{definition}

In training, we are given trajectories that may have (a) different initial conditions, and (b) different unknown ODE system parameters.
We observe a test trajectory (indexed by $M+1$) from time $t=t_0, \ldots, t_r$ and we wish to forecast its future after time $t_r$. The test trajectory can have an OOD initial condition \update{and OOD (unknown) ODE parameters $W^{(M+1)*}$.}

\begin{figure*}[t]
    \begin{subfigure}[t]{0.2\textwidth}
        \centering
        \includegraphics[scale=0.6]{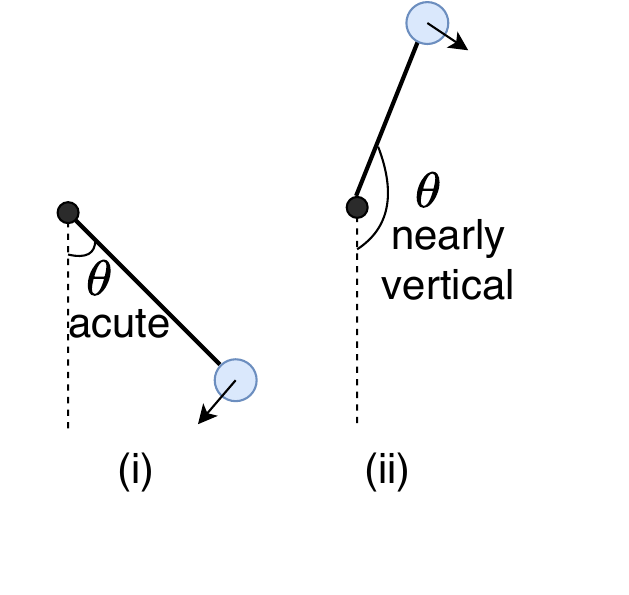}
        \vspace{-21pt}
        \caption{Pendulum task}
        \label{fig:pendulum_task_example_desc}
    \end{subfigure}
    ~~~~
        \begin{subfigure}[t]{0.37\textwidth}
            \centering
            \includegraphics[scale=0.3]{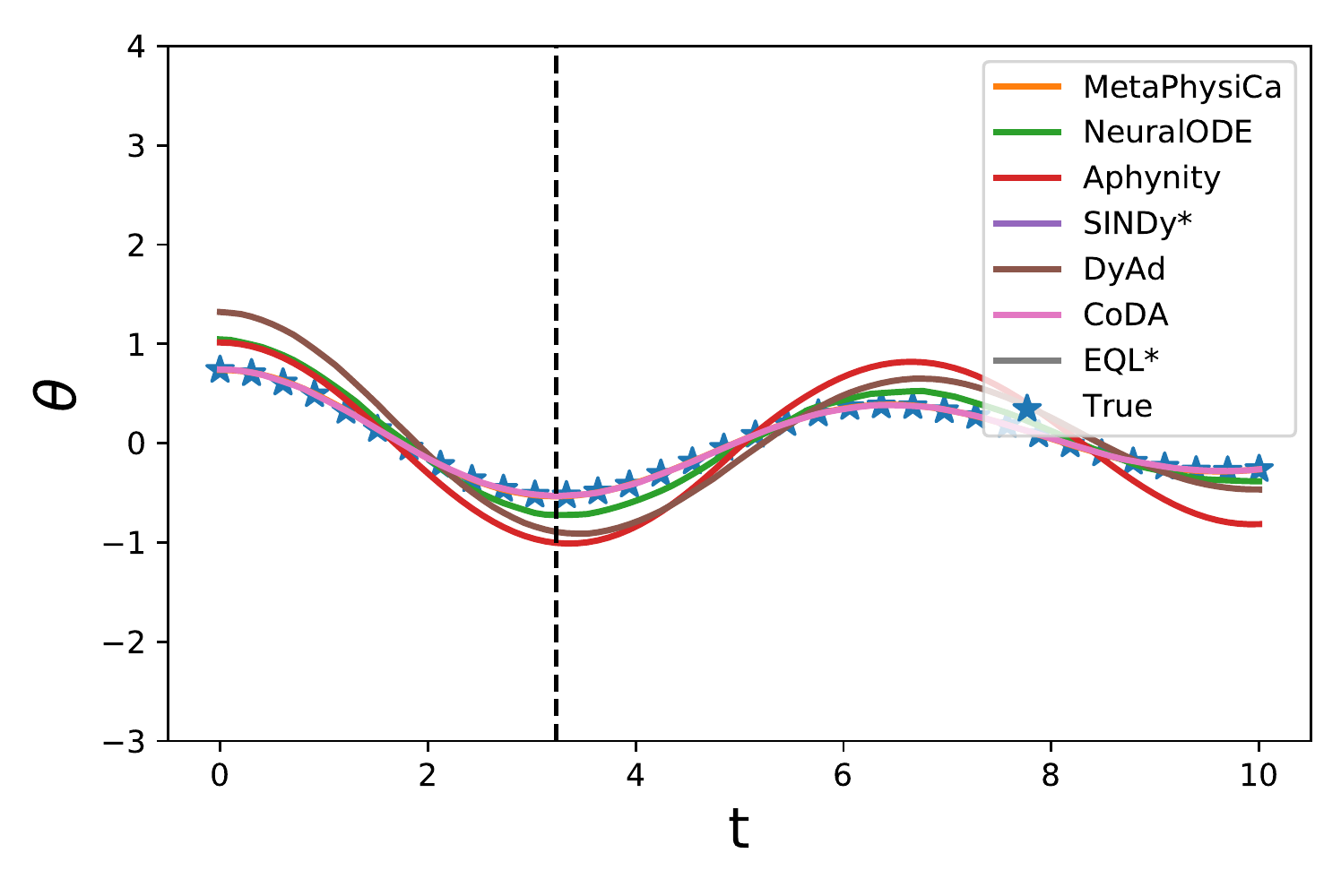}
            \vspace{-10pt}
            \caption{In-distribution predictions of $\theta_t$}
            \label{fig:damped_pendulum_results_id}
        \end{subfigure}
        ~~~~
        \begin{subfigure}[t]{0.37\textwidth}
            \centering
            \includegraphics[scale=0.3]{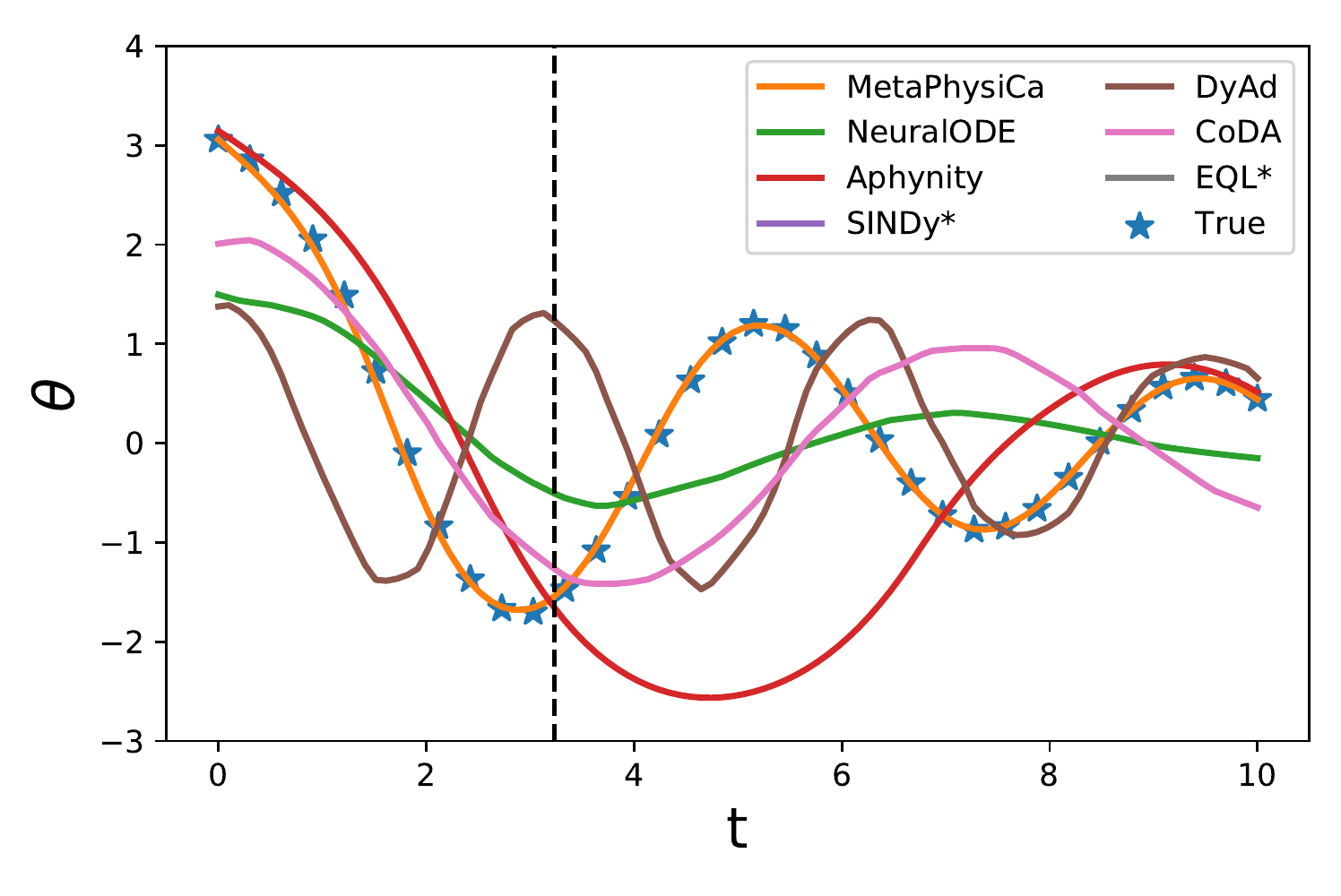}
            \vspace{-10pt}
            \caption{Predictions of $\theta_t$ under OOD $\xt{0}$}
            \label{fig:damped_pendulum_results_ood}
        \end{subfigure}
    
    \begin{subfigure}[t]{\textwidth}
        \centering
        \begin{tabular}{@{}lHHrrr@{}}
            \toprule
            & \multicolumn{4}{c}{Test NRMSE $\downarrow$} \\
            Methods & ID & OOD $\xt{0}$ & ID & OOD $\xt{0}$ & OOD $\xt{0}$ and $\mW^{*}$ \\
            \midrule
            \textbf{Standard Deep Learning} \\
            ~NeuralODE~\citep{chen2018} & 0.016 (0.008) & 0.739 (0.056) & 0.083 (0.033) & 0.591 (0.119) & 0.717 (0.210) \\
            \textbf{\update{Meta Learning}} \\
            ~DyAd~\citep{wang2021a} & 0.012 (0.000) & 0.695 (0.049) & 0.078 (0.051) & 0.834 (0.263) & 0.804 (0.267) \\
            ~CoDA~\citep{kirchmeyer2022} & 0.017 (0.011) & 0.669 (0.365) & 0.052 (0.032) & 0.764 (0.201) & 1.011 (0.226) \\
            \textbf{Physics-informed Machine Learning} \\
            ~APHYNITY~\citep{guen2020} & 0.047 (0.022) & 0.803 (0.168) & 0.097 (0.020) & 0.970 (0.384) & 1.159 (0.334) \\
            ~SINDy~\citep{Brunton2016} & $\nan$ & $\nan$ & $\nan$ & $\nan$ & $\nan$ \\
            ~EQL~\citep{martius2016} & $\nan$ & $\nan$ & $\nan$ & $\nan$ & $\nan$ \\
            ~\methodsindyvtwo {\bf (ours)} & 0.028 (0.005) & {\bf 0.078 (0.090)} & 0.049 (0.002) & {\bf 0.070 (0.011)} & {\bf 0.181 (0.012)}	\\
            \bottomrule
        \end{tabular}
    \caption{
    Normalized RMSE $\downarrow$ of test predictions from different methods in-distribution and two OOD scenarios. 
    $\nan$ indicates that the model returned errors during test-time predictions, for example, because the learnt ODE was too stiff (numerically unstable) to solve.}
    \label{tab:damped_pendulum_results}
    \end{subfigure}
\vspace{-5pt}
    \caption{\small
    \textbf{(a)} Predict pendulum motion from noisy observations: (i) in-distribution, when dropped from acute angles and (ii) OOD w.r.t initial conditions and parameters, when a different pendulum is dropped from nearly vertical angles.
    \textbf{(b, c)} shows example ground truth curves (blue stars) in- and out-of-distribution along with predictions from different models. While most tested methods perform well in-distribution, only \method (orange) closely follows the true curve OOD and all other methods are terribly non-robust.
    \textbf{(d)} Standard deep learning methods and physics-informed machine learning methods fail to forecast accurately out-of-distribution. 
    On the other hand, {\bf \method outputs up to $4\times$ more robust OOD predictions.} 
    }
    \label{fig:damped_pendulum_results}
\vspace{-10pt}
\end{figure*}

\vspace{-10pt}
\paragraph{Illustrative example.}
\Cref{fig:pendulum_task_example_desc} shows an example of an out-of-distribution task for forecasting the motion of a pendulum with friction. 
The state $\x_t = [\theta_t, \omega_t] \in \sR^2$ describes the angle made by the pendulum with the vertical and the corresponding angular velocity at time $t$.
The true (unknown) function $\psi$ describing this dynamical system is given by 
$\psi([\theta_t, \omega_t]; \thetapsi^*) = [\omega_t, -\alpha^{*2} \sin(\theta_t) - \rho^* \omega_t]$ with $\thetapsi^*=(\alpha^*, \rho^*)$ denoting the parameters relating to the pendulum's period and the damping coefficient.
\textbf{(1.)} In training, we observe $M$ (noisy) trajectories of motion over discrete time steps $t=0, 0.1, \ldots, 10$ from experiments (tasks) where a pendulum is dropped with no angular velocity.
Each training experiment is performed by dropping different pendulums (i.e., $\thetapsi^{(i)*} \sim P(\thetapsi^* | E=e^{(i)})$) from angles $0 < \theta_{t_0} < \pi/2$. 
\textbf{(2.)} In test, the experiment is repeated with a different distribution over the initial dropping angles, $\pi-0.1 < \theta_{t_0} < \pi$ (nearly vertical angles) and a different distribution over ODE parameters $\thetapsi^*$.
The test trajectory is observed over a smaller time window $t=0,0.1, \ldots, 3.3$ and the forecasting task is to predict the future states of the pendulum till time $t=10$.

\vspace{-7pt}
\section{Related work \& their limitations}
Next we describe different classes of existing approaches that are commonly used for the dynamical system forecasting and their inherent challenges out-of-distribution.

\vspace{-10pt}
\subsection{Neural network methods} \label{subsec:standard_neural_networks}
\vspace{-5pt}
Deep learning's ability to model complex phenomena has allowed it to make great strides in a number of physics applications~\citep{Lusch2018,Yeo2019,kochkov2021machine,dang2022tnt,brandstetter2022message}.
However, standard deep learning methods are known to learn spurious correlations and tend to fail when the test distribution of the inputs are different from that observed in training~\citep{wang2020bridging,geirhos2020}. 
\cref{fig:damped_pendulum_results} depicts the out-of-distribution failure of several deep learning methods from NeuralODE~\citep{chen2018} to more complex meta learning approaches~\citep{wang2021a} in our running damped pendulum example (more details of the experiment is in \cref{sec:results}). 
\update{While DyAd~\citep{wang2021a} and CoDA~\citep{kirchmeyer2022} use meta-learning objectives to adapt to new dynamical system parameters, they are not robust to OOD initial conditions.}

In standard deep learning tasks, \citet{xu2020} show that an MLP's failure to extrapolate to out-of-distribution can be traced to an absence of algorithmic alignment, which is an appropriate combination of basis and activation functions within the architecture for the task.
For example, the outputs of an MLP with ReLU activations will be linear far from the training domain even when trained to predict a sine/quadratic function.
For dynamical system forecasting, our \cref{fig:main_meta_physics_task}(c) depicts the results of a similar experiment for a standard sequence model (NeuralODE): the model can approximate the target sine function in the training domain (green region) but predicts a linear function far outside the training domain. 
This means that {\em we need algorithmic alignment (i.e., to include appropriate basis functions) in order to make accurate forecasts in OOD tasks.}

\vspace{-5pt}
\subsection{Physics-informed machine learning (PIML)} \label{subsec:piml}
To alleviate the challenges described above for standard neural networks, several physics-informed machine learning (PIML) methods have been proposed (e.g.,~\citep{willard2020,wang2020,faghmous2014big,Karpatnea}) that utilize physics-based domain knowledge about the dynamical system for better predictions. 
The type of physics-based knowledge vary across methods, for example, \textbf{(a)} a dictionary of basis functions (e.g., $\sin$, $\cos$, $\frac{d}{dt}$)~\citep{schmidt2009,Brunton2016,martius2016,Raissi2018,cranmer2020} related to the task, \textbf{(b)} a completely specified physics model~\citep{Raissi,Raissi2018,jiang2019} or with missing terms~\citep{guen2020}, and \textbf{(c)} different domain-specific physical constraints such as energy conservation~\citep{greydanus2019,cranmer2020lagrangian}, symmetries~\citep{wang2020b,finzi2021,brandstetter2022lie}.
While these PIML methods improve upon standard neural networks, \cref{fig:damped_pendulum_results} shows that they are generally not designed for OOD forecasting tasks. 
To precisely study the reasons for this failure, we categorize these methods into inductive and transductive methods based on requirements over the dynamical system parameters $\thetapsi^*$. 

\vspace{-10pt}
\paragraph{Transductive PIML methods.}
Transductive inference focuses on predicting missing parts from the training data. 
In PIML, transductive inference methods treat {\em training and test examples} as unrelated tasks, hence OOD generalization tends to be less of a challenge in transductive methods.
For instance, SINDy~\citep{Brunton2016}, EQL~\citep{martius2016}, and related methods \citep{Raissi2018,chen2021a}, learn the ODE equation based on a dictionary of basis functions for a specific parameter $\mW^{(i)*}$.
These transductive methods, however, do not transfer knowledge learnt in training to predicting test examples with a different $\mW^{(j)*}$. 
This forces these methods to forecast simply based on the initial observations of the test task alone, often leading to poor performance.  
\cref{fig:main_meta_physics_task}(d) illustrates this case where a transductive method (unsuccessfully) tries to learn the unknown parameter $\mW^{(3)*}$ of the test task from a few initial test observations.
Another class of transductive methods~\citep{Raissi,Raissi2017,yu2021a} assume that the ODE parameters $\thetapsi^*$ remain constant across all training and test tasks, and regularize neural networks to respect a given physics model.
Causal PINNs~\citep{wang2022respecting} further ensure that, for any time $t$, predictions at time less than $t$ are accurately resolved {\em before} predictions at time $t$. 
However, they do not allow for causal interventions to initial states and unknown parameters of the dynamical system. Further, these methods will perform poorly in-distribution if different training tasks have different ODE parameters.

\vspace{-8pt}
\paragraph{Inductive PIML methods.}
Taking the opposite approach, {\em inductive inference} focuses on learning rules from the training data that can be applied to unseen test examples. 
Inductive methods dominate PIML approaches but are fragile OOD, since the learned rules are learned within the scope of the training data and are not guarantee to work outside the training data scope.
For example, APHYNITY~\citep{guen2020} and NDS~\citep{mehta2021} are such inductive methods that augment a neural network to a known incomplete physics model where the parameters of the physics model are predicted inductively using a recurrent network. 
As illustrated in~\cref{fig:main_meta_physics_task}(e), these methods are able to learn from training tasks with different ODE parameters $\thetapsi^{(i)*}$. 
However, the recurrent network in APHYNITY fails OOD and often returns incorrect physics parameters OOD (see \Cref{fig:damped_pendulum_results_ood}). 
Further, the augmented neural network suffers from the same issues discussed in \cref{subsec:standard_neural_networks} leading to poor OOD performance as seen in \cref{fig:damped_pendulum_results}.

With these key reasons identified for the fragility of existing methods to OOD initial conditions, next we propose an approach ({\em \method}) that is more robust to these challenges and outputs more robust predictions out-of-distribution, while also giving accurate predictions in-distribution.

\vspace{-7pt}
\section{Proposed approach: \method} \label{sec:method}

\begin{figure}[t]
    \centering
    \vspace{-2pt}
    \includegraphics[scale=0.33]{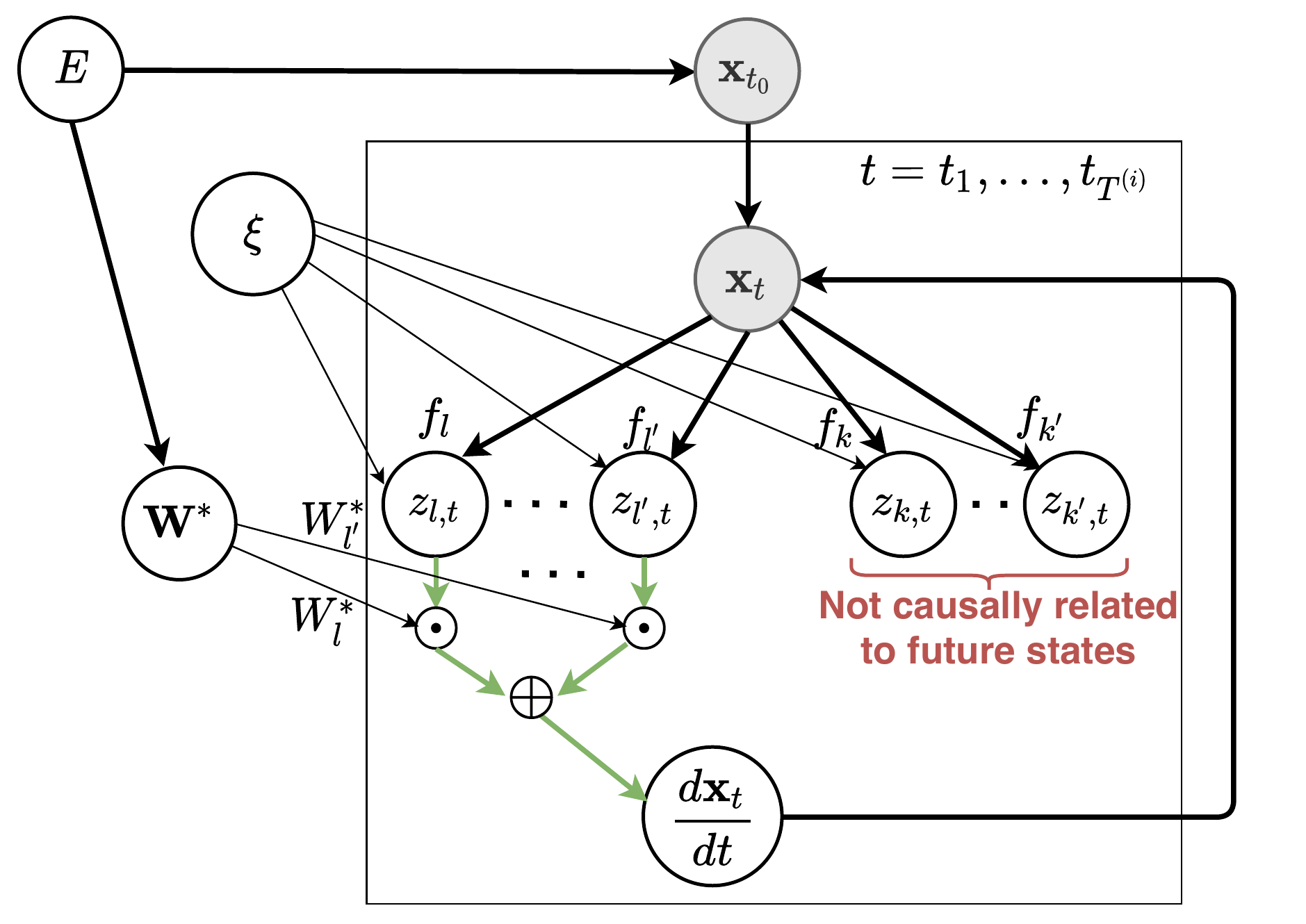}
    \caption{Deterministic SCM for a dynamical system. The dynamics is defined via an unknown linear combination of basis functions. \update{The distribution of initial conditions $\vx_{t_0}$ and ODE parameters $\thetapsi^*$ depends on the environment $E$.}}
    \label{fig:scm}
    \vspace{-15pt}
\end{figure}

We first describe a family of causal models, then explain how meta learning allows us to perform a hybrid transductive-inductive approach for improved OOD accuracy.

\vspace{-5pt}
\subsection{Structural causal model}
\vspace{-5pt}
We describe the dynamical system using a deterministic structural causal model~\citep{peters2020} with measurement noise over the observed states and explicitly define the assumptions over the unknown function $\psi$ in \cref{def:forecasting_task}. 

The causal diagram is depicted in \cref{fig:scm} in the plated notation iterating over time $t=t_0,\ldots,t_{T^{(i)}}$ for each task $\cT^{(i)}$. 
As before, the state of the dynamical system is $\x^{(i)}_t \in \sR^d$ for task $i$. 
We note that our SCM may not necessarily be the true SCM, but rather a SCM that is indistinguishable from the true one w.r.t.\ interventions to the environment variable $E$ that affects the initial conditions $\xt{0}^{(i)}$ and $\thetapsi^{(i)*}$. 
We define the causal process at each time step $t$ for $i$-th task as follows.

    Let $f_k(\cdot; \thetaglobal{k}):\sR^d \to \sR, 1\leq k \leq m$, be $m$ linearly independent basis functions each with a separate set of parameters $\thetaglobal{k}^*$ acting on an input state $\vx^{(i)}_t$. 
    Examples of such basis functions include trigonometric functions like $f_1(\vx^{(i)}_t; \thetaglobal{1}^*)=\sin(\thetaglobalelem{1}{1} x^{(i)}_{t, 1} + \thetaglobalelem{1}{2})$, polynomial functions like $f_2(\vx^{(i)}_t; \thetaglobal{2}) = x^{(i)}_{t, 1} x^{(i)}_{t, 2}$, and so on.
    The corresponding outputs from these basis are shown as $z^{(i)}_{k, t} := f_k(\vx^{(i)}_t; \thetaglobal{k})$ in \cref{fig:scm}.
    The derivative $\nicefrac{d\vx^{(i)}_{t, j}}{dt}$ for a particular dimension $j\in \{1,\ldots,d\}$ is only affected by a few (unknown) basis function outputs $z^{(i)}_{k, t}$ (green arrows in \cref{fig:scm}) and is a linear combination of these selected basis functions with coefficients $\thetapsi^{(i)*}$.
    However, these selected basis functions and their corresponding parameters $\thetaglobal{}$ are assumed to be invariant across all the tasks, i.e., $\nicefrac{d\vx^{(i)}_{t,j}}{dt}, j\in\{1,\ldots,d\}$, is defined using the same basis functions for all $i=1,\ldots,M$.
    Finally, the derivatives dictate the next state of the dynamical system.
    We observe the dynamical system with independent additive measurement noise $\x^{(i)}_t := \vx^{(i)}_t + {\bm \varepsilon}^{(i)}_t$, where ${\bm \varepsilon^{(i)}_t} \sim \mathcal{N}(\vzero, \sigma_\varepsilon^2 I)$.

We assume that we are given the collection of $m$ possible basis functions $f_k(\cdot; \thetaglobal{}), k=1,\ldots,m$, $m \geq 2$, with unknown $\thetaglobal{}$ and {\em no prior knowledge of which $\{f_k\}_{k=1}^m$ causally influence $\nicefrac{d\vx^{(i)}_{t}}{dt}$}. 
The need for basis functions stems from extensive experimentation and our analysis in \Cref{subsec:standard_neural_networks}, where we show that appropriate basis functions must be incorporated within the architecture in order to extrapolate to OOD scenarios (see \cref{fig:main_meta_physics_task}(c)).

\vspace{-5pt}
\subsection{Meta learning \& model architecture}
\vspace{-5pt}
Given the training data $\{(\vx^{(i)}_t)_t\}_{i=1}^M$ generated from the unknown SCM described above, our goal is three-fold: 
\textbf{(a)} discover the true underlying causal structure, i.e., which of the edges $z_{k,t} \to \nicefrac{d\vx_{t, j}}{dt}$ exist for $j=1,\ldots,d$, 
\textbf{(b)} learn the global parameters $\thetaglobal{}$ that parameterize the relevant basis functions, and
\textbf{(c)} learn the task-specific parameters $\thetapsi^{(i)*}$ that act as coefficients in linear combination of the selected basis functions.
In the following, we propose a meta-learning framework that introduces structure (gate) parameters $\Phi$ that are shared across tasks and task-specific coefficients $\mW^{(i)}$ that vary across the tasks
\vspace{-5pt}
\begin{align} \label{eq:model}
    \frac{d\xpred^{(i)}_t}{dt} = (\mW^{(i)} \odot \Phi) F(\xpred^{(i)}_t; \thetaglobal{}) \;,
\end{align}
where $\odot$ is the Hadamard product and
\vspace{-10pt}
\begin{itemize}[leftmargin=*,itemsep=0pt]
    \item $F(\xpred^{(i)}_t; \thetaglobal{}) := \begin{bmatrix} f_1(\xpred^{(i)}_t; \thetaglobal{1}) & \cdots & f_m(\xpred^{(i)}_t; \thetaglobal{m}) \end{bmatrix}^T$ is the vector of outputs from the basis functions with parameters $\thetaglobal{}$,
    \item $\Phi \in \{0, 1\}^{d \times m}$ are the learnable parameters governing the global causal structure across all tasks such that $\Phi_{j, k} = 1$  iff edge $z_{k,t} \to \nicefrac{d\vx_{t, j}}{dt}$ exists in \Cref{fig:scm},
    \item $\mW^{(i)} \in \sR^{d \times m}$ are task-specific parameters that act as coefficients in linear combination of the selected basis functions.
\end{itemize}
\vspace{-7pt}
Next we describe a procedure to obtain the structure parameters $\Phi$. 
Finding whether an edge exists or not in the causal graph is known as the causal structure discovery problem (e.g., ~\citet{heinze-deml2017}).
We use a score-based causal discovery approach (e.g., \cite{huang2018}) where we assign a score to each possible causal graph. 
We wish to find the {\em minimal} causal structure, i.e., with the least number of edges, that also fits the training data.
This balances the complexity of the causal structure with training likelihood, and avoids overfitting the training data.

A sparse structure for $\Phi$ implies fewer terms in the RHS of the learnt equation for the derivatives in \cref{eq:model}.  
Several causal discovery approaches have been proposed that learn such minimal causal structure via continuous optimization~\citep{zheng2018dags,ng2022masked}.
We use the log-likelihood of the training data with $\ell_1$-regularization term to induce sparsity that is known to perform well for general causal structure discovery tasks~\citep{zheng2018dags}.
Note that since the direction of all the edges are known (i.e., $z_{k,t} \to \nicefrac{d\vx_{t, j}}{dt}$), we do not need the acyclicity constraints and the causal graph is uniquely identified by its Markov equivalence class~\citep[Chapter 2]{pearl2009causality}.

The prediction error is given by ${R}^{(i)}(\mW^{(i)}, \Phi, \thetaglobal{}) := \frac{1}{T^{(i)} + 1} \sum_{t=t_0}^{t_{T^{(i)}}} ||\xpred^{(i)}_t - \x^{(i)}_t||_2^2$ where $\xpred^{(i)}_t = \xt{0}^{(i)} + \int_{t_0}^t (\mW^{(i)} \odot \Phi) F(\xpred^{(i)}_\tau; \thetaglobal{})  d\tau$ are the predictions obtained using an ODE solver to integrate \cref{eq:model}.
In practice however, we found the squared loss directly between the predicted and estimated ground truth derivatives, i.e., $\widetilde{R}^{(i)}(\mW^{(i)}, \Phi, \thetaglobal{}) = \frac{1}{T^{(i)}+1}\sum_{t=t_0}^{t_{T^{(i)}}} ||\nicefrac{d\xpred^{(i)}_t}{dt} - \nicefrac{d\x^{(i)}}{dt}||_2^2$, leads to a stable learning procedure with better accuracy in-distribution and OOD. 
As discussed before, we use an $\ell_1$-regularization term $||\Phi||_1$ to learn a causal structure with the fewest possible edges $z_{k, t} \to \nicefrac{d\vx_{t, j}}{dt}, j=1,\ldots,d,$ while minimizing the prediction error in training. 

Our structure discovery task comes with an additional challenge as the training tasks could have been obtained under different (hidden) environments (as defined in \cref{def:forecasting_task}).
While there are score-based (discrete optimization) approaches~\citep{ghassami2018multi,perry2022causal} for such non-IID data, aforementioned approaches based on continuous optimization (e.g.,~\citep{zheng2018dags}) are not guaranteed to learn the correct structure.
For example, they may output a structure that is optimal for one environment consisting of a large number of training tasks but suboptimal for other environments.

Our goal then is to learn a structure that minimizes the prediction error across all environments simultaneously, similar to learning robust representations via invariant risk minimization-type methods~\citep{arjovsky2019invariant,krueger2021out}.
Since the environment $e^{(i)}$ of a particular task $i$ is hidden to our approach, we use a modified V-REx regularization~\citep{krueger2021out} that minimizes the variance of prediction errors across tasks instead of environments, focusing on robustness to the worst-case scenario (that all tasks have unique environments).

Now we are ready to describe our final optimization objective. 
Similar to standard meta-learning objectives~\citep{finn2017,franceschi2018bilevel,hospedales2021meta}, we propose a 
bi-level objective that optimizes the structure parameters $\Phi$ and the global parameters $\xi$ in the outer-level, and the task-specific parameters $\mW^{(i)}$ in the inner-level as follows
\begin{align}\label{eq:loss}
\hat{\Phi}, \hat{\thetaglobal{}}
&= \argmin_{\Phi, \thetaglobal{}} \frac{1}{M} \sum_{i=1}^M R^{(i)}(\hat{\mW}^{(i)}, \Phi, \thetaglobal{}) + \lambdaone ||\Phi||_1 \nonumber \\ 
& \qquad \qquad + \lambdavrex \text{Variance}(\{R^{(i)}(\hat{\mW}^{(i)}, \Phi, \thetaglobal{})\}_{i=1}^M) \nonumber  \\
& \text{s.t.}~, \forall i, \hat{\mW}^{(i)} = \argmin_{\mW^{(i)}} R^{(i)}(\mW^{(i)}, \Phi, \thetaglobal{})  \:,
\end{align}
where $\lambdaone$ and $\lambdavrex$ are hyperparameters.
The bi-level optimization in \cref{eq:loss} can be approximated by alternate optimization steps for $(\Phi, \thetaglobal{})$ and $\{\mW^{(i)}\}_{i=1}^M$ in outer and inner loops respectively~\citep{borkar1997stochastic,chen2021closing}. 
In our experiments, jointly optimizing $\Phi, \thetaglobal{}$ and $\mW^{(i)}, i=1,\ldots,M,$ instead resulted in comparable performance with considerable computational benefits over alternating SGD.
The discrete structure parameters $\Phi$ can be approximated using (stochastic) Gumbel-Softmax variables~\citep{jang2016categorical,ng2022masked} or using deterministic binarization techniques~\citep{courbariaux2015binaryconnect,courbariaux2016binarized}. 
We use the latter and reparameterize $\Phi_{j, k}:=\vone(\sigma(\widetilde{\Phi}_{j,k}) > 0.5)$ where $\Phi' \in \sR^{d\times m}$, $\sigma(\cdot)$ is the sigmoid function, and the gradients are estimated via a straight-through-estimator.

{\em Hyperparameter selection:} We choose the hyperparameters $\lambdaone$ and $\lambdavrex$ that result in sparsest model (i.e., with the least $||\hat{\Phi}||_0$) while achieving validation loss within 5\% of the best validation loss in held-out {\em in-distribution} validation data. The use of {\em in-distribution data for validation} is key requirement since in OOD tasks one {\em does not} have access to samples from the test distribution.
Additional implementation details are provided in \cref{sec:appx_implementation_details}.

\method can be extended to more expressive structural causal models than \cref{fig:scm} that construct more expressive basis functions by composing them. 
\method with such an expressive SCM shows OOD performance gains on a complex ODE task (\cref{sec:appx_complex_ode_task}), but sometimes suffers from learning stiff ODEs due to the complexity of such a 2-layer composition procedure. 
Better optimization techniques may help alleviate this problem.

\vspace{-5pt}
\subsection{Transductive test-time adaptation}

Finally, given a test task $\widetilde{\mathcal{T}}^{(M+1)} = (\xt{0}^{(M+1)}, \ldots, \xt{r}^{(M+1)})$ 
with the unknown ground-truth parameters $\thetapsi^{(M+1)*} \sim P(\thetapsi^* | E=e^{(M+1)})$ as defined in \cref{def:forecasting_task},
we adapt the learnt model's task-specific parameters $\mW^{(M+1)}$ by optimizing the following while keeping $\hat{\Phi}, \hat{\thetaglobal{}}$ fixed
\begin{align}\label{eq:testadapt}
\hat{\mW}^{(M+1)} = \argmin_{\mW^{(M+1)}} &
\frac{1}{t_r + 1} \sum_{t=t_0}^{t_r} ||\xpred^{(M+1)}_t - \x^{(M+1)}_t||_2^2 
\end{align}
where $\xpred^{(M+1)}_t = \xt{0}^{(M+1)} + \int_{t_0}^t (\mW^{(M+1)} \odot \hat{\Phi}) F(\xpred^{(M+1)}_\tau; \hat{\thetaglobal{}})  d\tau$ are the predictions obtained using the optimal values $\hat{\Phi}, \hat{\xi}$.
Note the following two key aspects of the test-time adaptation in \cref{eq:testadapt}: \textbf{(a)} Only the task-specific parameters $\mW^{(M+1)}$ are adapted whereas the meta-model $\hat{\Phi}$ learnt during training is kept fixed, and \textbf{(b)} only the observations from time $t_0, \ldots, t_r$ of the given test trajectory is used to adapt the parameters $\mW^{(M+1)}$. 
Transductively adapting the task-specific parameters to the initial observations from the test trajectory allows the model to be robust to OOD ODE parameters $\thetapsi^{(M+1)*}$. 
The final predictions $(\xpred^{(M+1)}_t)_{t_r}^{t_{T^{(M+1)}}}$ from the model are obtained with the test-time adapted parameters $\hat{\mW}^{(M+1)}$ and the fixed parameters with no adaptation $\hat{\Phi}, \hat{\xi}$.

\cref{fig:metaphysica_architecture} in Appendix shows a schematic diagram of \method along with the training/test methodologies.

\section{Empirical evaluation} \label{sec:results}

\begin{figure*}
\centering
    \begin{subfigure}{\textwidth}
        \centering
        \begin{tabular}{@{}lHHrrr@{}}
            \toprule
            & \multicolumn{4}{c}{Test Normalized RMSE (NRMSE) $\downarrow$} \\
            Methods & ID & OOD & ID & OOD $\xt{0}$ & OOD $\xt{0}$ and $\mW^{*}$ \\
            \midrule
            \textbf{Standard Deep Learning} \\
            ~NeuralODE~\citep{chen2018} & 0.003 (0.000) & 1.229 (~~~~0.059) & 0.005 (0.000) & 1.139 (0.031) & 1.073 (0.102) \\
            \textbf{\update{Meta Learning}} \\
            ~DyAd~\citep{wang2021a} & & & 0.006 (0.001) & 1.147 (0.044) & 1.207 (0.202) \\ 
            ~CoDA~\citep{kirchmeyer2022} & 0.004 (0.001) & 2.044 (~~~~0.754) & 0.004 (0.001) & 1.341 (0.389) & 1.090 (0.274) \\
            \textbf{Physics-informed Machine Learning} \\
            ~APHYNITY~\citep{guen2020} & 0.075 (0.071) & 2.345 (~~~~6.035) & 0.151 (0.150) & 0.544 (0.249) & 0.898 (0.211) \\
            ~SINDy~\citep{Brunton2016} & 2.038 (0.008) & 2.447 (~~~~0.065) & 1.999 (0.046) & 2.746 (0.476) & $\nan$ \\
            ~EQL~\citep{martius2016} & $\nan$ & $\nan$ & $\nan$ & $\nan$ & $\nan$ \\
            ~\methodsindyvtwo {\bf (Ours)} & 0.006 (0.002) & {\bf 0.035 (~~~~0.028)} &        0.009 (0.004) & {\bf 0.019 (0.002)} & {\bf 0.100 (0.080)}\\
            \bottomrule
        \end{tabular}
        \caption{Test NRMSE  $\downarrow$ for different methods. $\nan$ indicates that the model returned errors during test.}
        \label{tab:epidemic_model_results}
    \end{subfigure}
    
    \begin{subfigure}[t!]{\textwidth}
        \centering
        \includegraphics[scale=0.2]{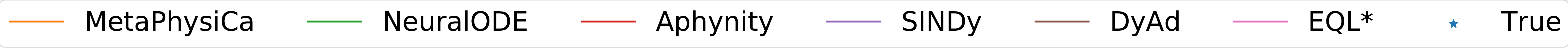}
    \end{subfigure}

    \begin{subfigure}{0.48\textwidth}
        \centering
        \includegraphics[scale=0.23]{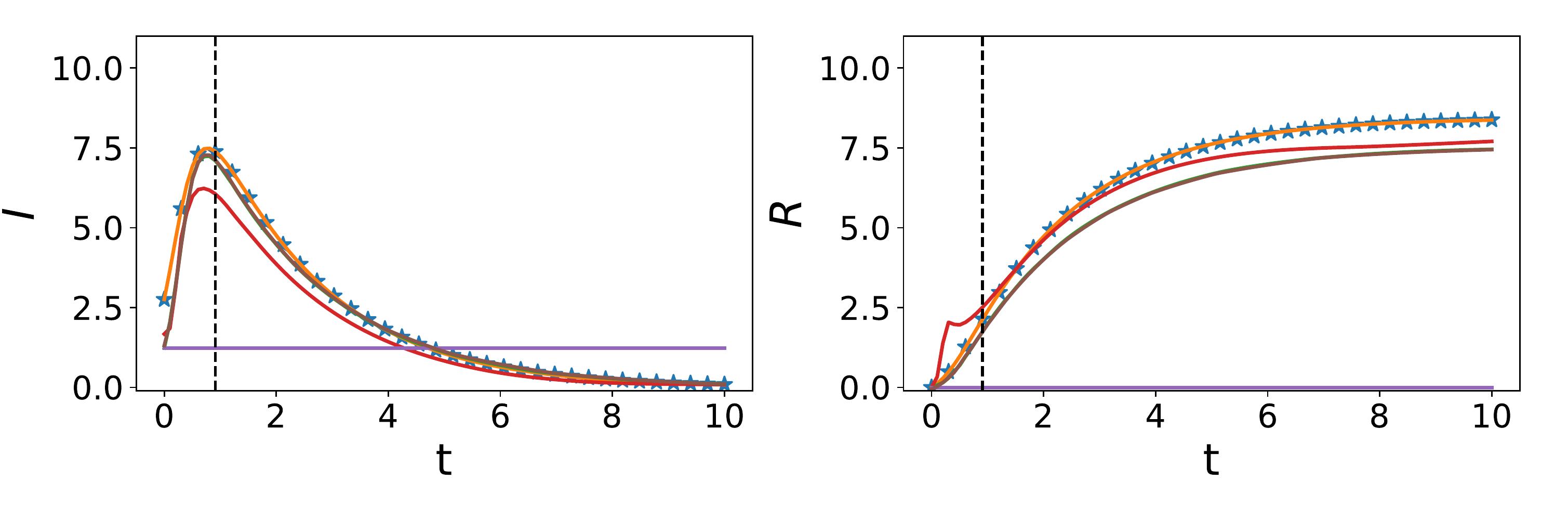}
        \vspace{-10pt}
        \caption{In-distribution predictions}
        \label{fig:epidemic_model_results_id}
    \end{subfigure}
    ~~
    \begin{subfigure}{0.48\textwidth}
        \centering
        \includegraphics[scale=0.23]{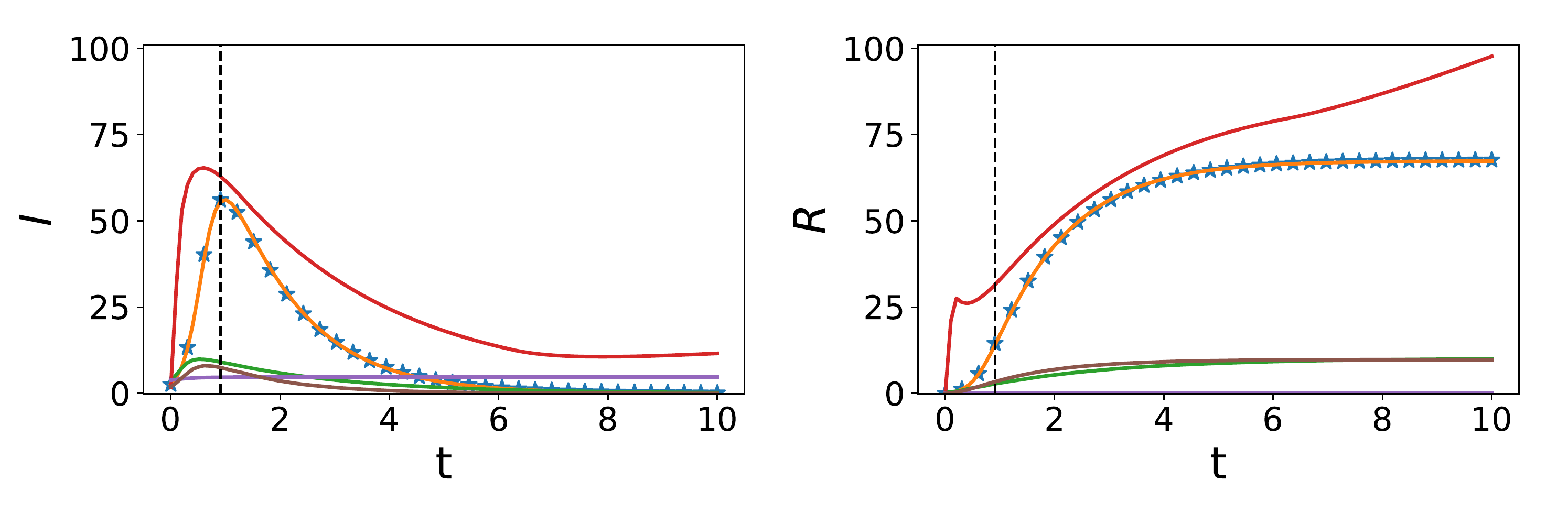}
        \vspace{-10pt}
        \caption{Predictions under OOD $\xt{0}$}
        \label{fig:epidemic_model_results_ood}
    \end{subfigure}
    \vspace{-5pt}
    \caption{\small
    \textbf{(Epidemic model results)} 
    \textbf{(a)} 
    {\bf \method outputs $28\times$ and $9\times$ more robust OOD predictions} for the two OOD scenarios respectively.
    \textbf{(b, c)} shows example ground truth curves (blue stars) in- and out-of-distribution along with corresponding predictions. Only \method (orange) closely follows the true curve OOD.
    }
    \label{fig:epidemic_model_results}
\vspace{-15pt}
\end{figure*}

We evaluate {\bf \method} in synthetic forecasting tasks based on 3 different dynamical systems (ODEs) from the literature~\citep{guen2020,wang2020bridging} adapted to our OOD scenario, namely, \textbf{(i)} Damped pendulum system, \textbf{(ii)} Predator-prey system and  \textbf{(iii)} Epidemic model. 
We compare against the following approaches: 
\textbf{(a) NeuralODE}~\citep{chen2018}, a deep learning method for learning ODEs, \textbf{(b) DyAd}~\citep{wang2021a} (modified for ODEs), 
that adapts to different training tasks with a weakly-supervised encoder, 
\textbf{(c) CoDA}~\citep{kirchmeyer2022}, that learns to modify its parameters to each environment with a low-rank adaptation, 
\textbf{(d) APHYNITY}~\citep{guen2020}, that augments a known incomplete physics model with a neural network, 
\textbf{(e) SINDy}~\citep{Brunton2016}, a {\em transductive} PIML method that uses sparse regression to learn linear coefficients over a given set of basis functions,
\textbf{(f) EQL}~\citep{martius2016}, a {\em transductive} PIML method that uses $\sin, \cos$ and other basis functions within a neural network and learns a sparse model.
Additional details about the models is presented in \cref{sec:appx_implementation_details}.

\vspace{-10pt}
\paragraph{Dataset generation.}

As per \cref{def:forecasting_task}, for each dynamical system, we simulate the respective ODE to generate $M=1000$ training tasks each observed over regularly-spaced discrete time steps $\{t_0, \ldots, t_{\T}\}$
where $\forall l, t_l = 0.1 l$.
For each training task $\cT^{(i)}, i = 1,\ldots,M$, we sample an initial condition $\xt{0}^{(i)} \sim P(\xt{0} | E=e)$ where $E=e$ is the training environment. 
Similarly, we sample different $\thetapsi^{(i)*} \sim P(\thetapsi^* | E=e)$ for each training task $i$.
At OOD test, we generate $M'=200$ test tasks by simulating the respective dynamical system over timesteps $\{t_0, \ldots, t_{\tobs}\}$, where again $\forall l, t_l = 0.1 l$.
\update{
For each test task $j = 1, \ldots, M'$, we sample test initial conditions $\xt{0}^{(j)} \sim P(\xt{0} | E=e')$ and test ODE parameters $\thetapsi^{(j)*} \sim P(\thetapsi^* | E=e')$, where $E=e'$ is the test environment. 
We consider two OOD scenarios: \textbf{(a)} \textbf{(OOD $\xt{0}$.)} when only the initial conditions are OOD, and \textbf{(b)} \textbf{(OOD $\xt{0}$ and $\thetapsi^{*}$.)} when initial conditions and ODE parameters are OOD. 
The latter can induce completely different test supports for both the initial conditions and the ODE parameters.
}

We consider three dynamical systems in our experiments, with 3 to 6 RHS terms in their respective differential equations: a {damped pendulum system}~\citep{guen2020}, a predator-prey system~\citep{wang2020bridging}, and an epidemic (SIR) model~\citep{wang2020bridging}, with following OOD shifts in their initial conditions respectively: acute initial angles in training to nearly vertical initial angles in OOD test, initial prey population $10\times$ less in OOD test than in training, and initial population susceptible to a disease $10\times$ more in OOD test than in training.
\update{For all three dynamical systems, all ODE parameters are $\approx 1.5\times$ higher OOD than in training (with non-overlapping support).}
We generate the damped pendulum dataset with 1\% zero-mean Gaussian noise and the rest with no noise to show that OOD failure of baselines is unrelated to noise: existing methods fail OOD even with clean observations.  
Detailed description of the datasets is presented in \cref{sec:appx_datasets}.

\paragraph{Results.}
We repeat our experiments 5 times with random seeds and report in-distribution (ID) and out-of-distribution (OOD) normalized root mean squared errors (NRMSE), i.e., RMSE normalized with standard deviation of the ground truth.
\cref{fig:damped_pendulum_results,fig:lotka_volterra_results,fig:epidemic_model_results} show the errors and example predictions from all models for the three datasets respectively.
The first column of Tables~\ref{tab:damped_pendulum_results}, \ref{tab:epidemic_model_results}, \ref{tab:lotka_volterra_results} shows in-distribution results while the last two columns show the respective OOD scenarios.
NeuralODE, DyAd, CoDA and APHYNITY use neural network components and are able to learn the in-distribution task well with low errors. 
However, the corresponding errors OOD are high as they are unable to adapt to OOD initial conditions and OOD parameters. 
Example OOD predictions (\cref{fig:damped_pendulum_results_ood,fig:lotka_volterra_results_ood,fig:epidemic_model_results_ood}) from these methods show that they have not learnt the true dynamics of the system. 
For example, for epidemic modeling (\cref{fig:epidemic_model_results_ood}), most models predict trajectories very similar to training trajectories even though the number of susceptible individuals is $10\times$ higher in OOD test. 
SINDy and EQL cannot use the training data and are fit on the test observations alone (see \cref{fig:main_meta_physics_task}(d)). 
Thus, they are unable to identify an accurate analytical equation from these few test observations, resulting in prediction issues due to stiff ODEs. 
\method performs the best OOD across all datasets achieving $2\times$ to $28\times$ lower NRMSE OOD errors than the best baseline.
\update{
\vspace{-10pt}
\paragraph{Qualitative analysis.}
\method's performance gains stem from two factors: \textbf{(i)} The optimal meta-model $\hat{\Phi}$ learns the ground truth ODE (possibly reparameterized) for all 3 dynamical systems (shown in \cref{sec:appx_analysis_ablation}), and \textbf{(ii)} the model adapts its task-specific parameters separately to each OOD test task.
The former is key for robustness over OOD initial states (via algorithmic alignment) and the latter helps to be robust over OOD parameters $\thetapsi^*$.  
We further show in an ablation study (\cref{sec:appx_actual_ablation}) that sparsity regularization (i.e., $||\Phi||_1$) and test-time adaptation (\cref{eq:testadapt}) are the most important components of \method; OOD performance degrades significantly without either. 
}

\vspace{-7pt}
\section{Conclusions}
\vspace{-3pt}
We considered the out-of-distribution task of forecasting a dynamical system (ODE) under new initial conditions and new ODE parameters.
We showed that existing PIML methods do not perform well in these tasks and proposed \method that uses a meta-learning framework to learn the causal structure for the shared dynamics across all environments, while adapting the task-specific parameters.
Results on three OOD forecasting tasks show that \method is more robust with $2\times$ to $28\times$ reduction in OOD error compared to the best baseline.
\textbf{Limitations \& future work:}
We believe that forecasting models should be robust to OOD shifts, and that our work takes a step in the right direction with several potential avenues for future research:
    \textbf{(i)} 
    Extending \method to forecasting PDEs under OOD scenarios is an interesting extension that requires an expanded set of basis functions that includes differential operators, and considering OOD boundary conditions.
    \textbf{(ii)} 
    Better optimization techniques to avoid learning stiff ODEs when extending \method to more expressive SCMs.

\section*{Acknowledgements}
This work was funded in part by the National Science Foundation (NSF) Awards CAREER IIS-1943364 and CCF-1918483, the Purdue Integrative Data Science Initiative, and the Wabash Heartland Innovation Network. Any opinions, findings and conclusions or recommendations expressed in this material are those of the authors and do not necessarily reflect the views of the sponsors.

\balance

\bibliographystyle{icml2023}

\newpage
\appendix
\onecolumn

\begin{center}
\Large Supplementary Material of ``\method: OOD Robustness in Physics-informed Machine Learning''
\end{center}

\section{Description of tasks} \label{sec:appx_datasets}
For each dynamical system, we simulate the respective ODE to generate $M=1000$ training tasks each observed over regularly-spaced discrete time steps $\{t_0, \ldots, t_{\T}\}$ where $\forall l, t_l = 0.1 l$.
Our data generation process is succinctly depicted in \cref{tab:datasets}.
For each dataset, the second column shows the state variables $\x_t$ and the unknown parameters $\thetapsi^*$. 
For each training task $\cT^{(i)}, i = 1,\ldots,M$, we sample an initial condition $\xt{0}^{(i)} \sim P(\xt{0} | E=e)$ where $E=e$ is the training environment (shown under ID columns of the table). 
We sample a different $\thetapsi^{(i)*} \sim \mathcal{U}(\mW_\param, 2\mW_\param)$ for each task $i$ with $\mW_\param$ shown in \cref{tab:datasets}.

At test, we generate $M'=200$ test tasks by simulating the respective dynamical system over timesteps $\{t_0, \ldots, t_{\tobs}\}$, where again $\forall l, t_l = 0.1 l$. 
For each test task $j = 1, \ldots, M'$, we sample initial conditions $\xt{0}^{(j)} \sim P(\xt{0} | E=e')$ where $E=e'$ is the test environment and can induce a completely different support for the initial conditions $\xt{0}^{(j)}$ than in training.
The distribution of the dynamical system parameters $\thetapsi^*$ is kept the same for ``OOD $\xt{0}$'' scenario but is shifted for ``OOD $\xt{0}$ and $\thetapsi^*$'' scenario.
In the latter, we sample a different $\thetapsi^{(j)*} \sim \mathcal{U}(2\mW_\param, 3\mW_\param)$ for each test task $j$ with $\mW_\param$ shown in \cref{tab:datasets}.

\paragraph{Damped pendulum system~\citep{guen2020}.} 
The state $\x_t = [\theta_t, \omega_t] \in \sR^2$ describes the angle made by the pendulum with the vertical and the corresponding angular velocity at time $t$.
The true (unknown) function $\psi$ describing this dynamical system is given by 
$\frac{d\theta_t}{dt} = \omega_t, \frac{d\omega_t}{dt} = -\alpha^{*2} \sin(\theta_t) - \rho^* \omega_t$ where $\thetapsi^*=(\alpha^*, \rho^*)$ are the dynamical system parameters. 
We simulate the ODE over time steps $\{t_0, \ldots, t_{\T}\}$ with $\forall l, t_l=0.1l, \T = 100$ in training and over time steps $\{t_0, \ldots, t_{\tobs}\}$ in test with $\tobs=\frac{1}{3}T$.
In training, the pendulum is dropped from initial angles $\theta^{(i)}_{t_0} \sim \mathcal{U}(0, \pi/2)$ with no angular velocity, whereas in OOD test, the pendulum is dropped from initial angles $\theta^{(j)}_{t_0} \sim \mathcal{U}(\pi - 0.1, \pi)$ and angular velocity $\omega^{(j)}_{t_0} \in \mathcal{U}(-1, 0)$.

\paragraph{Predator-prey system~\citep{wang2020bridging}.} 
We wish to model the dynamics between two species acting as prey and predator respectively. 
We adapt the experiment by \citet{wang2020bridging} to our out-of-distribution forecasting scenario according to \cref{def:forecasting_task}. 
Let $p$ and $q$ denote the prey and predator populations respectively. 
The ordinary differential equations describing the dynamical system is given by
$
\frac{dp}{dt} = \alpha^* p - \beta^* p q \:, 
\frac{dq}{dt} = \delta^* p q - \gamma^* q \:, 
$
where $\thetapsi^* = (\alpha^*, \beta^*, \gamma^*, \delta^*)$ are the (unknown) dynamical system parameters. 
We simulate the ODE over time steps $\{t_0, \ldots, t_{\T}\}$ with $\forall l, t_l=0.1l, \T = 100$ in training and over time steps $\{t_0, \ldots, t_{\tobs}\}$ in test with $\tobs=\frac{1}{3}T$.
We generate $M=1000$ training tasks with different initial prey and predator populations with prey $p_{t_0}^{(i)} \sim \mathcal{U}(1000, 2000)$ and predator $q_{t_0}^{(i)} \sim \mathcal{U}(10, 20)$ for each $i=1,\ldots,M$.
At OOD test, we generate $M'=200$ out-of-distribution (OOD) test tasks with different initial prey populations $p_{t_0}^{(j)} \sim \mathcal{U}(100, 200)$ but the same distribution for predator population $q_{t_0}^{(j)} \sim \mathcal{U}(10, 20)$. 

\paragraph{Epidemic modeling~\citep{wang2020bridging}.} 
We adapt the experiment by \citet{wang2020bridging} to our out-of-distribution forecasting scenario according to \cref{def:forecasting_task}. 
The state of the dynamical system is described by three variables: number of susceptible ($S$), infected ($I$) and recovered ($R$) individuals.
The dynamics is described using the following ODEs: $\frac{dS}{dt} = -\beta \frac{SI}{N}, \frac{dI}{dt} = \beta \frac{SI}{N} - \gamma I, \frac{dR}{dt} = \gamma I$, where $\thetapsi = (\beta, \gamma)$ are the (unknown) dynamical system parameters and $N=S+I+R$ is the total population. 
We simulate the ODE over time steps $\{t_0, \ldots, t_{\T}\}$ with $\forall l, t_l=0.1l, \T = 100$ in training and over time steps $\tobs=\frac{1}{10}T$.
We generate $M=1000$ training tasks with different initial populations for susceptible ($S$) and infected ($I$) individuals, while the number of initial recovered ($R$) individuals are always zero. 
In training, we sample $S^{(i)}_{t_0} \sim \mathcal{U}(9, 10)$ and $I^{(i)}_{t_0} \sim \mathcal{U}(1, 5)$ for each $i=1,\ldots,M$.  
At OOD test, we generate $M'=200$ out-of-distribution test tasks with a different initial susceptible population, $S^{(j)}_{t_0} \sim \mathcal{U}(90, 100)$, while keeping the same distribution for infected population.

\begin{figure*}[t]
    \begin{subfigure}{\textwidth}
        \centering
        \begin{tabular}{@{}lHHrrr@{}}
            \toprule
            & \multicolumn{4}{c}{Test Normalized RMSE (NRMSE) $\downarrow$} \\
            Methods & ID & OOD & ID & OOD $\xt{0}$ & OOD $\xt{0}$ and $\mW^{*}$ \\
            \midrule
            \textbf{Standard Deep Learning} \\
            ~NeuralODE~\citep{chen2018} & 0.012 (0.002) & 1.098 (0.108) & 0.193 (0.024) & 1.056 (0.141) & 0.969 (0.172) \\
            \textbf{\update{Meta Learning}} \\
            ~DyAd~\citep{wang2021a} &  0.016 (0.003) & 1.213 (0.216) & 0.244 (0.025) & 1.088 (0.373) & 1.025 (0.403) \\
            ~CoDA~\citep{kirchmeyer2022} & $\nan$ & $\nan$ & $\nan$ & $\nan$ & $\nan$ \\ 
            \textbf{Physics-informed Machine Learning} \\
            ~APHYNITY~\citep{guen2020}  & 0.047 (0.019) & 0.301 (0.139) & 0.421 (0.332) & 3.937 (1.686) & 1.281 (0.457) \\
            ~SINDy~\citep{Brunton2016} & $\nan$ & $\nan$ & $\nan$ & $\nan$ & $\nan$ \\
            ~EQL~\citep{martius2016} & $\nan$ & $\nan$ & $\nan$ & $\nan$ & $\nan$ \\
            ~\methodsindyvtwo {\bf (Ours)} & 0.008 (0.001) & {\bf 0.008 (0.000)} & 0.049 (0.008) & {\bf 0.129 (0.030)} & {\bf 0.434 (0.128)} \\
            \bottomrule
        \end{tabular}
        \caption{Test NRMSE $\downarrow$ for different methods. $\nan$ indicates that the model returned errors during test.}
        \label{tab:lotka_volterra_results}
    \end{subfigure}
    
\begin{subfigure}[t!]{\textwidth}
    \centering
    \includegraphics[scale=0.2]{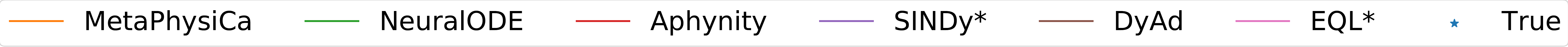}
\end{subfigure}

\begin{subfigure}{0.48\textwidth}
    \centering
    \includegraphics[scale=0.23]{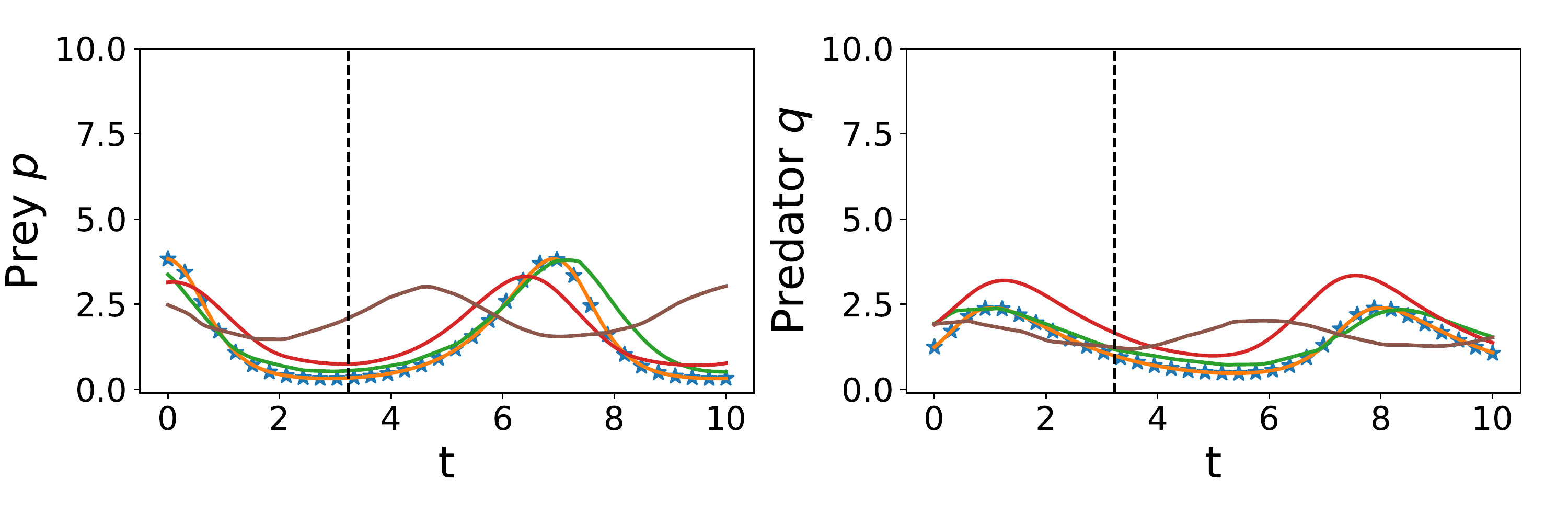}
    \caption{In-distribution predictions}
    \label{fig:lotka_volterra_results_id}
\end{subfigure}
~~
\begin{subfigure}{0.48\textwidth}
    \centering
    \includegraphics[scale=0.23]{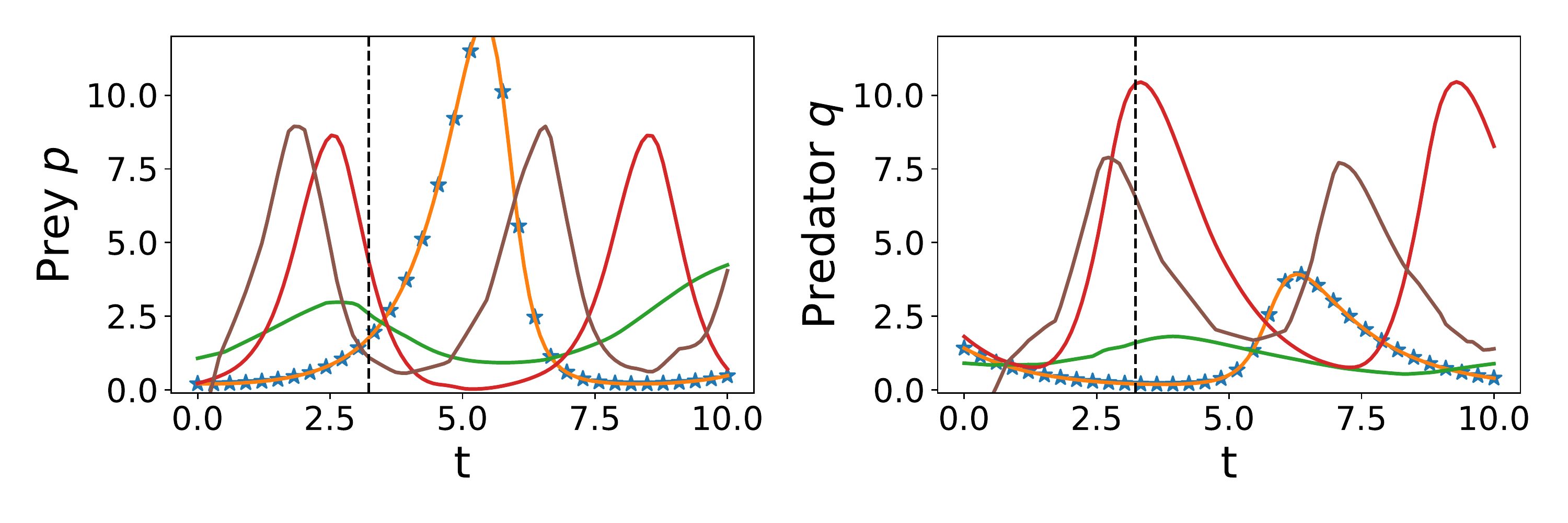}
    \caption{Predictions under OOD $\xt{0}$}
    \label{fig:lotka_volterra_results_ood}
\end{subfigure}
\vspace{-5pt}
\caption{\small
\textbf{(Predator-prey results)} 
\textbf{(a)} 
{\bf \method outputs $8\times$ and $2\times$ more robust OOD predictions} in the two OOD scenarios respectively.  
\textbf{(b, c)} shows example ground truth curves (blue stars) in- and out-of-distribution along with corresponding predictions. While most tested methods perform well in-distribution, only \method (orange) closely follows the true curve OOD.
}
\vspace{-15pt}
\label{fig:lotka_volterra_results}
\end{figure*}

\begin{table}[]
    \centering
    \begin{tabular}{@{}lllll@{}}
        \toprule
        Datasets & State variables & \multicolumn{1}{c}{ID} & \multicolumn{1}{c}{OOD $\xt{0}$} & \multicolumn{1}{c}{OOD $\xt{0}$ and $\thetapsi^*$} \\
        \midrule
        
        \multirow{3}{*}{Damped pendulum} & \multirow{2}{*}{$\x_t = (\theta_t, \omega_t)$} &  $\theta_0 \sim \cU(0, \pi/2)$ & $\theta_0 \sim \cU(\pi-0.1, \pi)$ & $\theta_0 \sim \cU(\pi-0.1, \pi)$ \\
        &&  $\omega_0 = 0$ & $\omega_0 \sim \cU(-1, 0)$ & $\omega_0 \sim \cU(-1, 0)$\\
            & $\thetapsi^* = (\alpha, \rho)$ & \multicolumn{3}{c}{$\alpha_\param = 1, \rho_\param = 0.2$} \\
            && \\
            \\
        
        \multirow{3}{*}{Predator prey system} & \multirow{2}{*}{$\x_t = (p_t, q_t)$} &  $p_0 \sim \cU(1000, 2000)$ & $p_0 \sim \cU(100, 200)$ & $p_0 \sim \cU(100, 200)$ \\
        && $q_0 \sim \cU(10, 20)$ & $q_0 \sim \cU(10, 20)$ & $q_0 \sim \cU(10, 20)$\\
            
            & $\thetapsi^* = (\alpha, \beta, \gamma, \delta)$ & \multicolumn{3}{c}{$\alpha_\param = 1, \beta_\param = 0.06, \gamma_\param = 0.5, \delta_\param = 0.0005$} \\
            && \\
            \\
            
        \multirow{4}{*}{Epidemic modeling} & \multirow{3}{*}{$\x_t = (S_t, I_t, R_t)$} &  $S_0 \sim \cU(9, 10)$ & $S_0 \sim \cU(90, 100)$ & $S_0 \sim \cU(90, 100)$ \\
        && $I_0 \sim \cU(1, 5)$ & $I_0 \sim \cU(1, 5)$ & $I_0 \sim \cU(1, 5)$\\
        && $R_0 = 0$ & $R_0 = 0$ & $R_0 = 0$\\
            
            & $\thetapsi^* = (\beta, \gamma)$ & \multicolumn{3}{c}{$\beta_\param = 4, \gamma_\param = 0.4$} \\
        \bottomrule
    \end{tabular}
    \caption{Description of the dataset generation process. For each dataset, $\x_t$ denotes the state variable of the dynamical system and $\mW^*$ denotes its parameters. 
    Column ``ID'' represents in-distribution initial states while the last two columns represent the two out-of-distribution scenarios.
    In-distribution ODE parameters $\thetapsi^{(i)*}$ are sampled from a uniform distribution $\thetapsi^{(i)*} \sim \mathcal{U}(\thetapsi_\param, 2\thetapsi_\param)$ and the out-of-distribution ODE parameters are sampled as $\thetapsi^{(i)*} \sim \mathcal{U}(2\thetapsi_\param, 3\thetapsi_\param)$.
    For example, in the damped pendulum dataset, in-distribution parameters are sampled as $\alpha^{(i)*} \sim \mathcal{U}(\alpha_\param, 2\alpha_\param) = (1, 2)$ and $\rho^{(i)*} \sim \mathcal{U}(\rho_\param, 2\rho_\param) = (0.2, 0.4)$ for each task $i$.  
    Similarly, the out-of-distribution ODE parameters (in the last column) are sampled as $\alpha^{(i)*} \sim \mathcal{U}(2\alpha_\param, 3\alpha_\param) = (2, 3)$ and $\rho^{(i)*} \sim \mathcal{U}(2\rho_\param, 3\rho_\param) = (0.4, 0.6)$.
    }
    \label{tab:datasets}
\end{table}

\section{Implementation details}\label{sec:appx_implementation_details}
In what follows, we describe implementation details of \method and the baselines.  

\subsection{\method}

\begin{figure}
    \centering
    \includegraphics[scale=0.3]{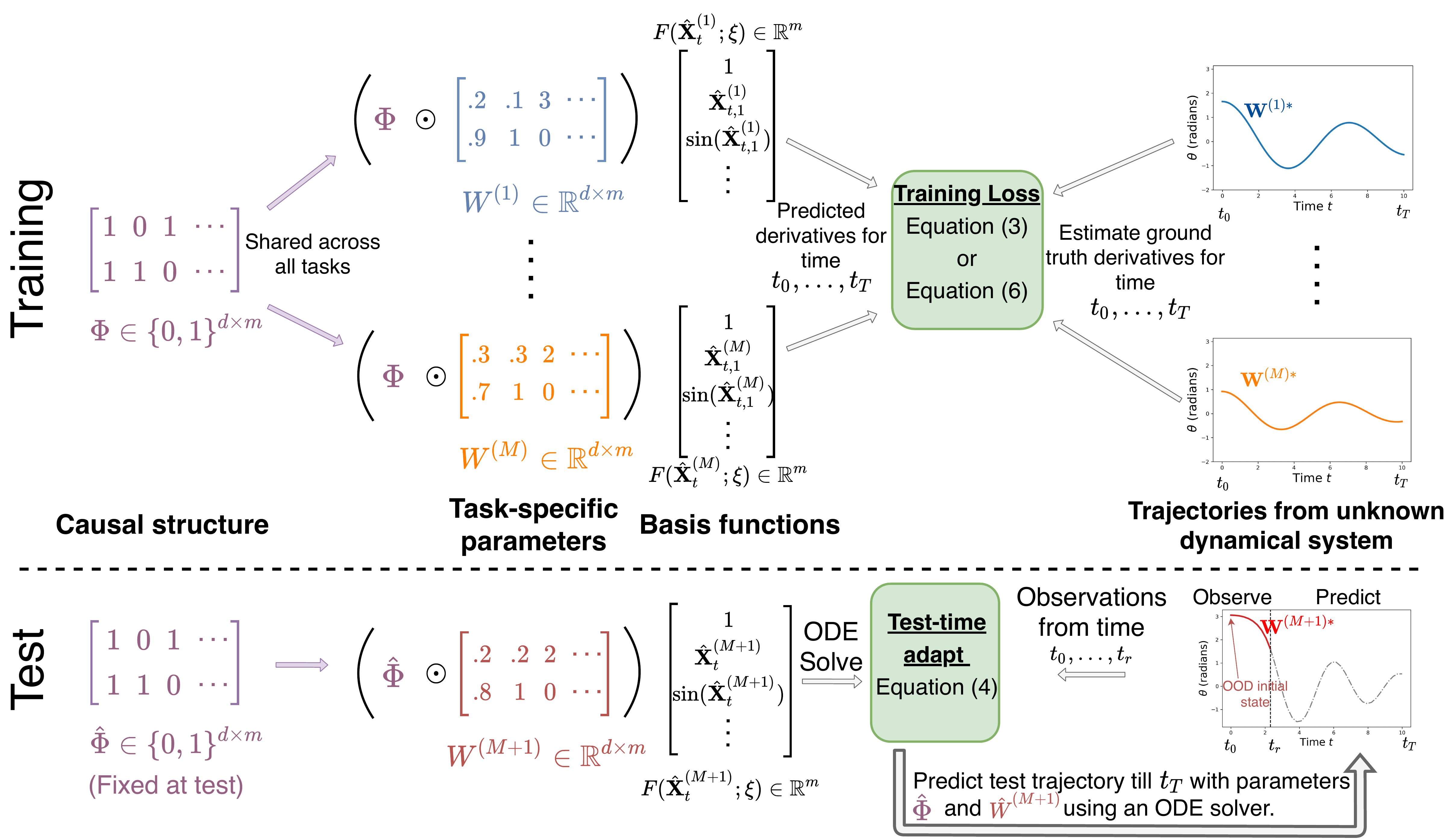}
    \caption{{Schematic diagram of \method and corresponding training/test methodologies. 
    We observe $M$ trajectories in training from the same dynamical system with different initial conditions and ODE parameters. 
    In training, $\Phi$, denoting the causal structure, is shared among all tasks $i=1,\ldots,M$, while $\mW^{(i)}$ are the task-specific parameters. 
    Predicted derivatives for task $i$ over time $t=t_0,\ldots,t_T$ are obtained from \cref{eq:model} using the parameters $\Phi,\mW^{(i)}$ and the basis functions $F(\x^{(i)}_t; \thetaglobal{})$. 
    During test, we adapt $\mW^{(M+1)}$ over the observations of the test trajectory from time $t_0, \ldots, t_r$, keeping the learnt causal structure $\hat{\Phi}$ fixed. 
    }}
    \label{fig:metaphysica_architecture}
\end{figure}

\cref{fig:metaphysica_architecture} shows a schematic diagram of \method and the corresponding training/test procedures. 
Recall from \cref{eq:model} that the proposed model is defined as
\begin{align}
    \frac{d\xpred^{(i)}_t}{dt} = (\mW^{(i)} \odot \Phi) F(\xpred^{(i)}_t; \thetaglobal{}) \;,
\end{align}
where $\odot$ is the Hadamard product and
\begin{itemize}[leftmargin=*,itemsep=0pt]
    \item $F(\xpred^{(i)}_t; \thetaglobal{}) := \begin{bmatrix} f_1(\xpred^{(i)}_t; \thetaglobal{1}) & \cdots & f_m(\xpred^{(i)}_t; \thetaglobal{m}) \end{bmatrix}^T$ is the vector of outputs from the basis functions with parameters $\thetaglobal{}$,
    \item $\Phi \in \{0, 1\}^{d \times m}$ are the learnable parameters governing the global causal structure across all tasks such that $\Phi_{j, k} = 1$  iff edge $z_{k,t} \to \nicefrac{d\vx_{t, j}}{dt}$ exists,
    \item $\mW^{(i)} \in \sR^{d \times m}$ are task-specific parameters that act as coefficients in linear combination of the selected basis functions.
\end{itemize}

In our experiments, we use polynomial and trigonometric basis functions, such that  
\begin{align*}
 F(&\xpred^{(i)}_t; \thetaglobal{}) := \\
 &\begin{bmatrix} 
 1 &
 \underbrace{\xpred^{(i)}_{t, 1} \ldots  \xpred^{(i)}_{t, d}}_{\text{polynomial order 1}} & 
 \underbrace{\xpred^{(i)2}_{t, 1} \ldots \xpred^{(i)}_{t, l-1} \xpred^{(i)}_{t, l} \ldots \xpred^{(i)2}_{t, d}}_{\text{polynomial order 2}} & 
 \underbrace{\sin(\xi_{1,1} \xpred^{(i)}_{t, 1} + \xi_{1, 2}) \ldots \sin(\xi_{d, 1} \xpred^{(i)}_{t, d} + \xi_{d, 2})}_{\text{trigonometric}} 
 \end{bmatrix}^T \:.
\end{align*} 

\cref{eq:loss} describes a bi-level objective that optimizes the structure parameters $\Phi$ and the global parameters $\xi$ in the outer-level, and the task-specific parameters $\mW^{(i)}$ in the inner-level as follows
\begin{align*}
\hat{\Phi}, \hat{\thetaglobal{}}
&= \argmin_{\Phi, \thetaglobal{}} \frac{1}{M} \sum_{i=1}^M R^{(i)}(\hat{\mW}^{(i)}, \Phi, \thetaglobal{}) + \lambdaone ||\Phi||_1 \nonumber +  \lambdavrex \text{Variance}(\{R^{(i)}(\hat{\mW}^{(i)}, \Phi, \thetaglobal{})\}_{i=1}^M) \nonumber  \\
&\qquad \text{s.t.}~ \hat{\mW}^{(i)} = \argmin_{\mW^{(i)}} R^{(i)}(\mW^{(i)}, \Phi, \thetaglobal{})  \quad \forall i=1,\ldots,M \:, 
\end{align*}
where $\lambdaone$ and $\lambdavrex$ are hyperparameters.
As discussed in the main text, the jointly optimizing $\Phi, \thetaglobal{}$ and $\mW^{(i)}, i=1,\ldots,M,$ instead of alternating SGD resulted in comparable performance with considerable computational benefits.
We use the following joint optimization objective to approximate \cref{eq:loss},
\begin{align}
\hat{\Phi}, \hat{\thetaglobal{}}, \hat{\mW}^{(1)}, \ldots, \hat{\mW}^{(M)}
= \argmin_{\Phi, \thetaglobal{}, {\mW}^{(1)}, \ldots, {\mW}^{(M)}} 
&\frac{1}{M} \sum_{i=1}^M R^{(i)}({\mW}^{(i)}, \Phi, \thetaglobal{}) + \lambdaone ||\Phi||_1  \\ \nonumber 
& + \lambdavrex \text{Variance}(\{R^{(i)}({\mW}^{(i)}, \Phi, \thetaglobal{})\}_{i=1}^M) 
\end{align}

We perform a grid search over the following hyperparameters: regularization strengths $\lambdaone \in \{10^{-4}, 10^{-3}, 5\times 10^{-3}, 10^{-2}\},
\lambdavrex \in \{0, 10^{-3}, 10^{-2}\}$, and learning rates $\eta \in \{10^{-2}, 10^{-3}, 10^{-4}\}$.
We choose the hyperparameters that result in sparsest model (i.e., with the least $||\hat{\Phi}||_0$) while achieving validation loss within 5\% of the best validation loss in held-out {\em in-distribution} validation data.

\subsection{NeuralODE~\citep{chen2018}}
The prediction dynamics corresponding to the latent NeuralODE model is given by
$\frac{d\xpred_t}{dt} = F_\nn(\xpred_t, \vz_{\leq \tobs}; \mW_1)$ where 
$\vz_{\leq \tobs} = F_\text{enc}(\xt{0}, \ldots, \xt{\tobs}; \mW_2)$ encodes the initial observations using a recurrent neural network $F_\text{enc}$ (e.g., GRU), and $F_\nn$ is a feedforward neural network. 
The model is trained with an ODE solver (dopri5) and the gradients computed using the adjoint method~\citep{chen2018}.
We perform a grid search over the following hyperparameters: number of layers for $F_\nn$, $L \in \{1,2,3\}$, size of each hidden layer of $F_\nn$, $d_h \in \{32, 64, 128\}$, size of the encoder representation $\vz_{\leq \tobs}$, $d_z \in \{32, 64, 128\}$, batch sizes $B \in \{32, 64\}$, and learning rates $\eta \in \{10^{-2}, 10^{-3}, 10^{-4}\}$.

\subsection{DyAd (modified for ODEs)~\citep{wang2021a}}
DyAd, originally proposed for forecasting PDEs, uses a meta-learning framework to adapt to different training tasks by learning a per-task weak label. 
We modify their approach for our ODE-based experiments. 
Since we do not assume the presence of weak labels for supervision for adaptation, we use mean of each variable in the training task as the task's weak label. 
We use NeuralODE as the base sequence model for the forecaster network. 
The forecaster network takes the initial observations as input and forecasts the future observations while being adapted with the encoder network. 
The encoder network is a recurrent network {(GRU in our experiments)} that takes as input the initial observations and predicts the weak label. 
The last layer representation from the encoder network is used to adapt NeuralODE via AdaIN~\citep{huang2017arbitrary}. 
{We perform a grid search over the following hyperparameters: size of hidden layers for the forecaster and encoder networks $d_h \in \{32, 64, 128\}$, number of layers for the forecaster network, $L \in \{1,2,3\}$, batch sizes $B \in \{32, 64\}$, and learning rates $\eta \in \{10^{-2}, 10^{-3}, 10^{-4}\}$.}

\subsection{APHYNITY~\citep{guen2020}}
APHYNITY assumes that we are given a (possibly incomplete) physics model $\phi(\cdot, \Theta_\phy)$ with parameters $\Theta_\phy$. 
When the training data may consist of tasks with different $\thetapsi^{(i)}*$, APHYNITY predicts the physics parameters with respect to the task $i$ inductively using a recurrent neural network $G_\nn$ from the initial observations of the system as $\hat{\Theta}^{(i)}_\phy = G_\nn(\xt{0}, \ldots, \xt{\tobs}; \mW_2)$. 
Then, APHYNITY augments the given physics model $\phi$ with a feedforward neural network component $F_\nn$ and defines the final dynamics as $\frac{d\xpred^{(i)}_t}{dt} = \phi(\xpred^{(i)}_t; \hat{\Theta}^{(i)}_\phy) + F_\nn(\xpred^{(i)}_t; \mW_1)$.
APHYNITY solves a constrained optimization problem to minimize the norm of the neural network component while still predicting the training trajectories accurately.
The model is trained with an ODE solver (dopri5) and the gradients computed using the adjoint method~\citep{chen2018}.
In our experiments, we provide APHYNITY with simpler physics models:
\begin{itemize}[leftmargin=*]
    \item For damped pendulum system, we use a physics model that assumes no friction: $\frac{d\theta_t}{dt} = \omega_t, \frac{d\omega_t}{dt} = -\alpha_\phy^{2} \sin(\theta_t)$ where $\Theta_\phy = \alpha_\phy$ is the physics model parameter.
    \item For predator-prey system, we use a physics model that assumes no interaction between the two species: 
    $\frac{dp}{dt} = \alpha_\phy p \:, \frac{dq}{dt} = - \gamma_\phy q$ where $\Theta_\phy = (\alpha_\phy, \gamma_\phy)$ are the physics model parameters.
    \item For epidemic model, we use a physics model that assumes the disease is not infectious:
    $\frac{dS}{dt} = 0, \frac{dI}{dt} = -\gamma I, \frac{dR}{dt} = \gamma I$, where $\Theta_\phy = \gamma_\phy$ is the physics model parameter.
\end{itemize}
In each dataset, APHYNITY needs to augment the physics model with a neural network component for accurate predictions.

{We perform a grid search over the following hyperparameters: number of layers for $F_\nn$, $L \in \{1,2,3\}$, size of each hidden layer of $F_\nn$, $d_h \in \{32, 64, 128\}$, batch sizes $B \in \{32, 64\}$, and learning rates $\eta \in \{10^{-2}, 10^{-3}, 10^{-4}\}$.}

\subsection{SINDy~\citep{Brunton2016}}
SINDy uses a given dictionary of basis functions to model the dynamics as
$\frac{d\xpred_t}{dt} = \Theta(\xpred_t) \mW$ where $\Theta$ is feature map with the basis functions (such as polynomial and trigonometric functions) and $\mW$ is simply a weight matrix. 
SINDy is trained using sequential threshold least squares (STLS) for sparse weights $\mW$. 
{We perform a grid search over the following hyperparameters: threshold parameter used in STLS optimization, $\tau_0 \in \{0.005, 0.01, 0.05, 0.1, 0.2, 0.5\}$, and the regularization strength $\alpha \in \{0.05, 0.01, 0.1, 0.5\}$.}

\subsection{Equation Learner~\citep{martius2016}}
Equation learner (EQL) is a neural network architecture where each layer is defined as follows with input $\vx$ and output $\vo$
\begin{align*}
\vz &= \mW\vx + \vb \\
\vo &= (f_1(z_1), f_2(z_2), \ldots, g_1(z_k, z_{k+1}), g_2(z_{k+2}, z_{k+3}), \ldots,) \:,
\end{align*}
where $f_i$ are unary basis functions (such as $\sin$, $\cos$, etc.) and $g_i$ are binary basis functions (such as multiplication). 
We use $\text{id}, \sin$ and multiplication functions in our implementation.
EQL is trained using a sparsity inducing $\ell_1$-regularization with hard thresholding for the final few epochs. 
{We perform a grid search over the following hyperparameters: number of EQL layers, $L\in \{1,2\}$, number of nodes for each type of basis function, $h \in \{1, 3, 5\}$, regularization strength $\alpha \in \{10^{-1}, 10^{-2}, 10^{-3}, 10^{-4}, 10^{-5}\}$, batch sizes $B \in \{32, 64\}$, and learning rates $\eta \in \{10^{-2}, 10^{-3}, 10^{-4}\}$.}

\section{Additional results} \label{sec:appx_additional_results}

\subsection{Qualitative analysis} \label{sec:appx_analysis_ablation}
{
Recall from \cref{eq:model} that the proposed model is defined as
\begin{align}
    \frac{d\xpred^{(i)}_t}{dt} = (\mW^{(i)} \odot \Phi) F(\xpred^{(i)}_t; \thetaglobal{}) \;,
\end{align}
where
$F(\xpred^{(i)}_t; \thetaglobal{})$ is the vector of outputs from the basis functions, $\Phi \in \{0, 1\}^{d \times m}$ are the learnable parameters governing the global causal structure across all tasks, and $\mW^{(i)} \in \sR^{d \times m}$ are task-specific parameters that act as coefficients in linear combination of the selected basis functions.

After training, the ODE learnt by the model can be easily inferred by checking all the terms in $\Phi$ that are greater than zero, i.e., $\Phi_{j, k} > 0$ implies $f_k(\vx_t; \thetaglobal{k}) \to \nicefrac{d\vx_{t, j}}{dt}$ exists in the causal graph. 
In other words, RHS of learnt ODE for $\nicefrac{d\vx_{t, j}}{dt}$ contains the basis function $f_k(\vx_t; \thetaglobal{k})$. 

\cref{tab:learnt_equations} shows the ground truth ODE and the learnt ODE for the three experiments. 
For each learnt ODE, we also depict the learnable parameters $W_l$ that can be adapted using \cref{eq:testadapt} during test-time. 
For damped pendulum and predator-prey system, the RHS terms in the learnt ODE exactly matches ground truth ODE, and from \cref{fig:damped_pendulum_results,fig:lotka_volterra_results}, it is clear that the method is able to accurately adapt the learnable parameters $W_l$ during test-time.
For epidemic modeling task, \method learns a reparameterized version of the ground truth ODE. 
For example, \method learns $\frac{dR_t}{dt} = W'_a I_t S_t + W'_b I^2_t + W'_c I_t R_t$, which can be written as $\frac{dR_t}{dt} = W_a I_t$ (the ground truth ODE) if $W'_a = W'_b = W'_c$, because $S_t + I_t + R_t = N$ is a constant denoting the total population.
While the learnt reparameterized ODE is more complex because it allows different values for $W'_a, W'_b, W'_c$, the test-time adaptation of these learnable parameters with the initial test observations results in them taking the same values.
}

\begin{table}[]
    \centering
    \resizebox{\textwidth}{!}{
    {
    \begin{tabular}{@{}llll@{}}
        \toprule
        Datasets & State variables & Ground truth ODE & Learnt ODE (from $\Phi$)\\
        \midrule
        
        \multirow{2}{*}{Damped pendulum} & \multirow{2}{*}{$\x_t = (\theta_t, \omega_t)$} & $\frac{d\theta_t}{dt} = \omega_t$ & $\frac{d\theta_t}{dt} = W_1 \omega_t$ \\
        && $\frac{d\omega_t}{dt} = -\alpha^{*2} \sin(\theta_t) - \rho^* \omega_t$ & $\frac{d\omega_t}{dt} = W_2 \sin(\theta_t) + W_3 \omega_t$ \\
            
        \\
        \multirow{2}{*}{Predator prey system} & \multirow{2}{*}{$\x_t = (p_t, q_t)$} & $\frac{dp_t}{dt} = \alpha^* p_t - \beta^* p_t q_t$ & $\frac{dp_t}{dt} = W_1 p_t + W_2 p_t q_t$ \\
        && $\frac{dq_t}{dt} = \delta^* p_t q_t - \gamma^* q_t$ & $\frac{dq_t}{dt} = W_3 p_t q_t + W_4 q_t$ \\
            
        \\
        \multirow{3}{*}{Epidemic modeling} & \multirow{3}{*}{$\x_t = (S_t, I_t, R_t)$} & $\frac{dS_t}{dt} = -\beta^* \frac{S_t I_t}{S_t + I_t + R_t}$ & $\frac{dS_t}{dt} = W_1 S_t I_t$ \\
        && $\frac{dI_t}{dt} = \beta^* \frac{S_t I_t}{S_t + I_t + R_t} - \gamma^* I_t$ & $\frac{dI_t}{dt} = W_2 S_t I_t + W_3 I_t^2 + W_4 I_t R_t$ \\
        && $\frac{dR_t}{dt} = \gamma^* I_t$ &  $\frac{dR_t}{dt} = W_5 S_t I_t + W_6 I_t^2 + W_7 I_t R_t$ \\
        
        \bottomrule
    \end{tabular}
    }
    }
    \caption{
        {
        \textbf{(Qualitative analysis.)} Ground truth dynamical system vs learnt ODE in the meta-model $\Phi$. Recall that $\Phi \in \{0, 1\}^{d \times m}$ dictates which of the basis functions affect the output $\nicefrac{d\x_t}{dt}$. The weights $W_l$ in the learnt ODE column are learnable parameters that are optimized via test-time adaptation in \cref{eq:testadapt}. \textbf{\method learns the exact ground truth ODE for Damped pendulum and Predator-prey system, and a reparameterized version of the true ODE for epidemic modeling task.}
        }
    }
    \label{tab:learnt_equations}
\end{table}

\subsection{Ablation results} \label{sec:appx_actual_ablation}
{
We present an ablation study comparing different components of \method in \cref{tab:ablation_results}. 
Table shows out-of-distribution test NRMSE for \method without each individual component on the three dynamical systems (OOD w.r.t $\xt{0}$).  
We observe that sparsity regularization (i.e., $||\Phi||_1$) and test-time adaptation are the most important components.  
For two out of three tasks, the method returns prediction errors without sparsity regularization. 

When testing \method without test-time adaptation, we simply use the mean of the task-specific weights learnt for training tasks as the task-specific weight for the given test trajectory, i.e., $\hat{\mW}^{M+1} = \frac{1}{M} \sum_i \mW^{(i)}$. 
This results in high OOD errors showing the importance of test-time adaptation.
V-REx penalty~\citep{krueger2021out} helps in some experiments and performs comparably in others. 
}

\begin{table}[]
    \centering
    {
        \begin{tabular}{@{}lrrr@{}}
            \toprule
            & \multicolumn{3}{c}{Test Normalized RMSE $\downarrow$ (OOD $\xt{0}$)} \\
            Method & Damped Pendulum & Predator-Prey & Epidemic Modeling \\
            \midrule 
            \methodsindyvtwo & {\bf 0.070 (0.011)} &  {\bf 0.129 (0.030)} &  {\bf 0.019 (0.002)} \\
            ~~ without $||\Phi||_1$ & $\nan$ & 1.806 (0.736) &  $\nan$ \\
            ~~ without test-time adaptation & 1.223 (0.741) & 1.404 (3.794) & 0.358 (0.554)	 \\
            ~~ without V-REx penalty & {\bf 0.070 (0.014)} & {\bf 0.129 (0.030)} & 0.042 (0.065) \\
            \bottomrule
        \end{tabular}
        \caption{
        {
            \textbf{(Ablation.)}  
            Out-of-distribution test NRMSE for \method without each individual component on the three dynamical systems (OOD w.r.t. $\xt{0}$ alone).  
            \textbf{Sparsity regularization (i.e., $||\Phi||_1$) and test-time adaptation are the most important components, whereas the V-REx penalty~\citep{krueger2021out} helps in some tasks, and performs comparably in others.}
        }
        }
        \label{tab:ablation_results}
    }
\end{table}

\subsection{Complex ODE Task} \label{sec:appx_complex_ode_task}

{

In this section, we extend \method to consider significantly more expressive structural causal models (compared to \cref{fig:scm}) that allow for composition of the basis functions. 
This is achieved with a 2-layer learnable basis function composition procedure. 
For example, given basis functions $f_1(\vx_t; \xi_1) = \sin(\xi_{1,1} \vx_{t,1} + \xi_{1,2})$, and $f_2(\vx_t; \xi_2) = \vx_{t,1} \vx_{t,2}$, one can construct more expressive basis functions with compositions: $\tilde{f}_3(\vx_t;\xi_3)=\sin(\xi_{3,3} \sin(\xi_{3, 1} \mathbf{x}_{t, 1} + \xi_{3,2}) + \xi_{3,4})$, $\tilde{f}_4(\vx_t;\xi_4)= \vx_{t,1} \vx_{t, 2} \sin(\xi_{4,1} \vx_{t, 1} + \xi_{4,2})$, etc., where $\xi_j$ are global parameters that remain constant for all training/test tasks. 
The rest of the SCM remains the same and the derivative $\nicefrac{d\mathbf{x}^{(i)}_{t, j}}{dt}$ for a particular dimension $j\in \{1,\ldots,d\}$ is a sparse linear combination of the original basis functions and the more expressive second layer ones. 

We evaluated MetaPhysiCa on a more complex ODE task from \citet{chen2020learning} adapted to our setting. 
We consider a two-dimensional ODE with state $\x_t = [p_t, q_t] \in \sR^2$: $\frac{dp_t}{dt} = a^* \sin(p_t) + b^* \sin(q_t^2); \frac{dp_t}{dt} = c^* \sin(p_t)\cos(q_t)$, where $\thetapsi^*=(a^*, b^*, c^*)$ are the dynamical system parameters. 
We simulate the ODE over time steps $\{t_0, \ldots, t_{\T}\}$ with $\forall l, t_l=0.1l, \T = 100$ in training and over time steps $\{t_0, \ldots, t_{\tobs}\}$ in test with $\tobs=\frac{1}{3}T$.
In training, we sample initial states $p_t, q_t \sim \mathcal{U}(0.5, 1)$, whereas in out-of-distribution test, we sample $p_t, q_t \sim \mathcal{U}(1, 1.5)$. 
The dynamical system parameters are sampled as $a^{(i)*}, b^{(i)*}, c^{(i)*} \sim \mathcal{U}(1.0, 1.5)$.

\cref{tab:complex_toy_results} shows the results for this task.  
First, we note that due to the complexity of a 2-layer learnable basis function procedure, we sometimes need to use validation data (held out from training) to cross-validate the learned model (and reject meta-models that do not do well in validation). 
\method learnt a stiff ODE for 2 out of 5 folds of cross-validation, resulting in no predictions for in-distribution validation data, which were rejected (marked as superscript $*$).  
In these experiments MetaPhysiCa performs $1.5\times$ to $1.7\times$ better than the competing baselines. 
We believe there is room for improvement in the optimization procedure of these more complex models.
}

\begin{table}[]
    \centering
    {
        \begin{tabular}{@{}lHHrr@{}}
            \toprule
            & \multicolumn{4}{c}{Test Normalized RMSE (NRMSE) $\downarrow$} \\
            Methods & ID & OOD & ID & OOD $\xt{0}$\\
            \midrule
            ~NeuralODE~\citep{chen2018} &  0.012 (0.001)~ & 0.188 (0.025)~ & 0.034 (0.008)~ &  0.296 (0.064)~ \\
            ~APHYNITY~\citep{guen2020}  & 0.010 (0.002)~ & 0.329 (0.050)~ & 0.027 (0.010)~ & 0.684 (0.117)~ \\
            ~SINDy~\citep{Brunton2016} & $\nan$ & $\nan$ & $\nan$ & $\nan$ \\
            ~\methodsindyvtwo (Ours) & 0.119 (0.072)\textsuperscript{*} & 0.110 (0.048)\textsuperscript{*} & 0.188 (0.035)\textsuperscript{*} & 0.203 (0.046)\textsuperscript{*} \\
            \bottomrule
        \end{tabular}
        }
        \caption{
        {Test NRMSE $\downarrow$ for different methods. 
        \textsuperscript{*} indicates that the method returned errors during predictions due to learning a stiff ODE.
        }       
        }
        \label{tab:complex_toy_results}
\end{table}

\end{document}